\documentclass[]{uiuc_paper}
\pdfoutput=1
\usepackage[toc,page,header]{appendix}
\usepackage{cleveref}
\usepackage{subcaption}
\usepackage{booktabs}
\usepackage{tabularx}

\usepackage{latexsym}
\usepackage{url}
\usepackage{amssymb}
\usepackage{amsmath}
\usepackage{amsfonts}
\usepackage[utf8]{inputenc}
\usepackage{booktabs}
\usepackage{pifont}
\usepackage{multirow}
\usepackage{makecell}
\usepackage{paralist}
\usepackage{xspace}
\usepackage{adjustbox}

\usepackage{colortbl}
\usepackage{algorithm}
\usepackage[noend]{algpseudocode}

\crefname{section}{Sec.}{Secs.}\Crefname{section}{Sec.}{Secs.}
\crefname{subsection}{Sec.}{Secs.}\Crefname{subsection}{Sec.}{Secs.}
\crefname{subsubsection}{Sec.}{Secs.}\Crefname{subsubsection}{Sec.}{Secs.}
\crefname{appendix}{App.}{Apps.}\Crefname{appendix}{App.}{Apps.}
\crefname{subappendix}{App.}{Apps.}\Crefname{subappendix}{App.}{Apps.}
\crefname{subsubappendix}{App.}{Apps.}\Crefname{subsubappendix}{App.}{Apps.}
\crefname{figure}{Fig.}{Figs.}\Crefname{figure}{Fig.}{Figs.}
\crefname{table}{Tab.}{Tabs.}\Crefname{table}{Tab.}{Tabs.}
\crefname{equation}{Eq.}{Eqs.}\Crefname{equation}{Eq.}{Eqs.}
\crefname{algorithm}{Alg.}{Algs.}\Crefname{algorithm}{Alg.}{Algs.}

\definecolor{oursbg}{HTML}{E9E9F4}
\newcommand{\ours}{\rowcolor{oursbg}}
\newcommand{\nJevCalls}{88,480}
\newcommand{\jevCostPerCall}{\$$6.5{\times}10^{-5}$}
\newcommand{\jevLatency}{8.4}
\newcommand{\adaptExtraPct}{42}
\newcommand{\adaptCalls}{3.1}
\newcommand{\adaptDollars}{\$0.0002}
\newcommand{\pruneCalls}{14}
\newcommand{\pruneDollars}{\$0.0009}
\newcommand{\oracleOverall}{0.90}
\newcommand{\pspEightIR}{0.75}
\newcommand{\pspEightJev}{0.82}
\newcommand{\pspEightNfe}{152}
\newcommand{\pruneJev}{0.82}
\newcommand{\adaptJevNfe}{73}
\newcommand{\nSched}{eleven}
\newcommand{\nAdapt}{252}
\newcommand{\adaptAboveAll}{all 252}
\newcommand{\allocGapJev}{61 [56, 65]}
\newcommand{\allocGapHps}{16 [11, 21]}
\newcommand{\allocGapIr}{25 [19, 30]}
\newcommand{\allocDiffHps}{45 [39, 51]}
\newcommand{\allocDiffIr}{36 [30, 42]}
\newcommand{\qualWinAll}{198}
\newcommand{\hpsAbove}{eight}
\newcommand{\irAbove}{one}

\newcommand{\genCells}{8}
\newcommand{\genPoolAdapt}{+0.050}
\newcommand{\genPoolPrune}{+0.046}
\newcommand{\genPoolAdaptCI}{[+0.042, +0.057]}
\newcommand{\genPoolPruneCI}{[+0.038, +0.053]}
\newcommand{\genGbPoolAdaptCI}{[+0.004, +0.016]}
\newcommand{\genGbPoolPruneCI}{[+0.008, +0.019]}
\newcommand{\genGbCells}{4}
\newcommand{\genGbPoolAdapt}{+0.010}
\newcommand{\genGbPoolPrune}{+0.013}
\newcommand{\genGbAbove}{above}

\newcommand{\qwAdapt}{0.784}
\newcommand{\qwAdaptNfe}{71}
\newcommand{\qwPrune}{0.808}
\newcommand{\qwOne}{0.678}
\newcommand{\qwR}{16}
\newcommand{\qwK}{6}
\newcommand{\qwTau}{0.996}
\newcommand{\qwPrompts}{553}
\newcommand{\qwImages}{2{,}212}

\newcommand{\ablPairCompHol}{+0.031 [+0.011, +0.052]}
\newcommand{\ablPairHolHps}{+0.094 [+0.059, +0.129]}
\newcommand{\ablPairHolIr}{+0.103 [+0.077, +0.135]}
\newcommand{\ablPruneHolHps}{+0.051 [+0.038, +0.065]}
\newcommand{\ablPruneHolIr}{+0.063 [+0.048, +0.079]}
\newcommand{\ablPosPrune}{+0.060 [+0.026, +0.095]}
\newcommand{\ablPosBonEight}{+0.049 [+0.022, +0.077]}

\newcommand{\scalePruneSixteenNfe}{288}
\newcommand{\scalePruneThirtyTwoNfe}{576}
\newcommand{\scaleBonSixteenNfe}{640}
\newcommand{\scaleBonThirtyTwoNfe}{1280}
\newcommand{\scalePruneSixteenMin}{+0.048 [+0.029, +0.069]}
\newcommand{\scalePruneSixteenMinJudge}{VQAScore}
\newcommand{\scalePruneThirtyTwoMin}{+0.048 [+0.019, +0.078]}
\newcommand{\scalePruneThirtyTwoMinJudge}{VQAScore}
\newcommand{\scaleBonMin}{+0.055 [+0.026, +0.085]}
\newcommand{\scaleBonMinJudge}{VQAScore}

\newcommand{\seqTwoTau}{0.7}
\newcommand{\seqTwoK}{6}
\newcommand{\seqTwoNfe}{72}
\newcommand{\seqTwoVsAdapt}{+0.019 [+0.008, +0.030]}

\newcommand{\compPrompts}{553}
\newcommand{\compExact}{546}
\newcommand{\compExactPct}{98.7}
\newcommand{\compObjectPct}{99.6}
\newcommand{\compCountPct}{100.0}
\newcommand{\compColourPct}{99.7}
\newcommand{\compPositionPct}{99.0}

\newcommand{\readgenSd}{+0.037 [+0.003, +0.067]}
\newcommand{\readgenFlux}{+0.063 [+0.033, +0.099]}
\newcommand{\readgenSdxl}{+0.114 [+0.081, +0.153]}
\newcommand{\readgenSigModels}{SD3.5-Medium, FLUX.1-dev and SDXL}

\newcommand{\dropinPspEightIr}{+0.071 [+0.055, +0.086]}
\newcommand{\dropinPspTwelveIr}{+0.069 [+0.054, +0.086]}
\newcommand{\dropinBonMinLo}{+0.013}

\newcommand{\fkAcc}{0.709}
\newcommand{\fkVsPrune}{+0.110 [+0.091, +0.128]}
\newcommand{\fkVsAdapt}{+0.089 [+0.072, +0.107]}

\newcommand{\fkjAcc}{0.754}
\newcommand{\fkjVsPrune}{+0.065 [+0.048, +0.082]}
\newcommand{\fkjVsAdapt}{+0.044 [+0.027, +0.061]}
\newcommand{\fkjOverFk}{+0.045 [+0.029, +0.061]}

\newcommand{\hpsvPruneVsBonEight}{+0.045 [+0.032, +0.059]}
\newcommand{\hpsvAdaptVsBonEight}{+0.024 [+0.010, +0.038]}
\newcommand{\hpsvBonMinLo}{+0.015}
\newcommand{\hpsvAdaptNfe}{73}

\newcommand{\sameprocMinLo}{+0.007}
\newcommand{\sameprocAll}{54}
\newcommand{\robustSameMinLo}{+0.026}
\newcommand{\robustSameAll}{22}
\newcommand{\robustSchedules}{11}
\newcommand{\scaleSameKMinLo}{+0.019}

\newcommand{\readgenPrevSig}{32}
\newcommand{\readgenPrevAll}{32}
\newcommand{\readgenPrevMinLo}{+0.001}
\newcommand{\supTestColVqa}{+0.092 [+0.033, +0.143]}
\newcommand{\supTestCountVqaCross}{+0.055 [+0.023, +0.087]}

\newcommand{\method}{Keystone\xspace}

\hypersetup{
  bookmarksnumbered=true,
  pdftitle={Keystone: Judging Previews at Key Steps for Test-Time Search in Diffusion Models},
  pdfauthor={Jiajun Fan, Chaoran Cheng, Ziwen Wang, Changyi Yang, Bangji Yang, Qinglong Zheng, Ge Liu}}

\title{Keystone: Judging Previews at Key Steps\\for Test-Time Search in Diffusion Models}

\author[1]{Jiajun Fan}
\author[1]{Chaoran Cheng}
\author[1,2]{Ziwen Wang}
\author[3]{Changyi Yang}
\author[1]{Bangji Yang}
\author[1]{Qinglong Zheng}
\author[1]{Ge Liu}

\affiliation[1]{\affillogo{\includegraphics{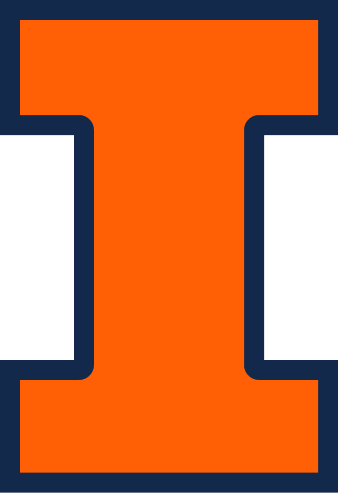}}{UIUC}}
\affiliation[2]{\affillogo{\includegraphics{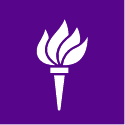}}{NYU}}
\affiliation[3]{\affillogo{\includegraphics{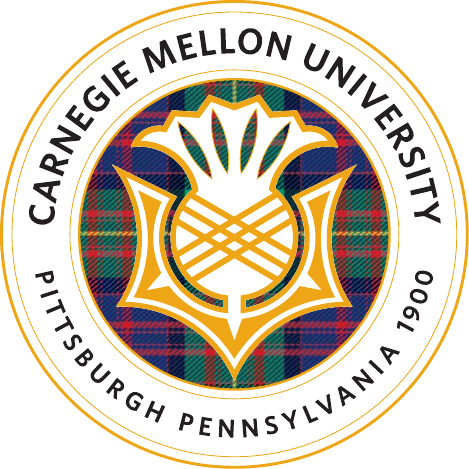}}{CMU}}

\abstract{Text-to-image diffusion and flow models yield a preview, their clean-image estimate, at every denoising step.
When they miscount, drop or misplace objects, test-time search is in a bind: its verifiers are reward models built for finished images, so judging early misreads blurry previews, and judging late, as \mbox{best-of-$N$} does, pays to fully denoise doomed candidates.
We find that the image is not the problem: the outcome is settled as denoising starts, by the seed or by a stochastic sampler's first steps, and already shows in early previews; what fails on them is the judge, as reward models read them near chance on several kinds of prompts.
We introduce \method, a training-free judge of previews that compiles a prompt into typed questions about its objects, counts, colours and positions and has an off-the-shelf multimodal decision model answer them.
Asked this way, the judge reads counts and positions from early previews better than reward models read them from finished images.
Searching at these key steps on GenEval with SD3.5-Medium, \method's adaptive search and pruning significantly outperform best-of-8 with every other verifier at under half its generator compute, and as a drop-in verifier there, the \method score significantly improves best-of-$N$, progressive seed pruning and FK steering; without retuning, both searches stay significantly above the strongest other verifier's best-of-$N$ on average over four base models and three benchmarks.
\method thus moves test-time judgment to the key steps where the outcome already shows.
}

\date{\today}

\begin{document}

\maketitle

\section{Introduction}
\label{sec:intro}

\begin{figure}[t]
\centering
\includegraphics[width=0.55\linewidth]{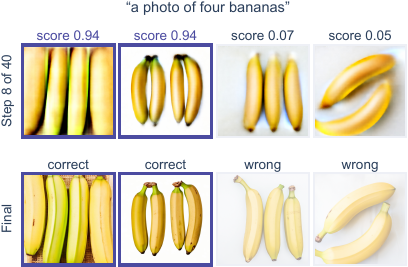}
\caption{\textbf{The outcome is visible long before denoising ends.}
Four of eight candidates for a development prompt on SD3.5-Medium.
\method asks whether there are exactly four bananas and scores each candidate's preview at step 8 of 40 (top); of the four shown, its pruning keeps the two it rates highest (framed) and need not finish the other two, which the official evaluator marks wrong (bottom, finished for reference). Example chosen by a fixed rule (\cref{sec:app_teaser}).}
\label{fig:teaser}
\end{figure}

\begin{figure}[t]
\centering
\includegraphics[width=\linewidth,trim=0.0 1.0 0.0 4.0,clip]{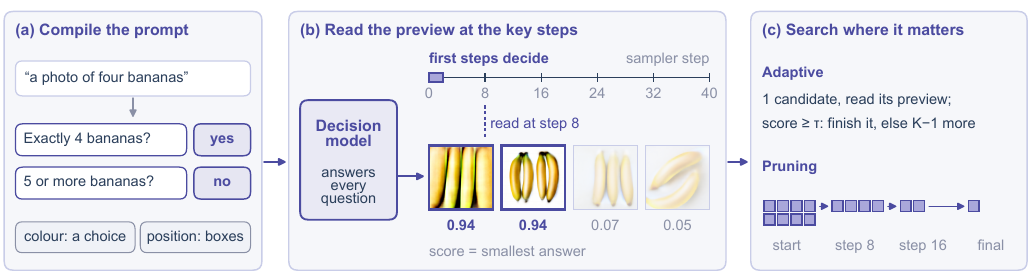}
\caption{\textbf{\method compiles the prompt, reads previews at the key steps and searches there.} (a) The prompt becomes typed questions about its slots, and an image's score is its smallest slot score (\cref{sec:questions}). (b) The outcome is decided in the first steps and read at a key step, here step 8 of the example of \cref{fig:teaser} (\cref{sec:keysteps,sec:exp_keysteps}). (c) Adaptive search and pruning spend compute by the preview scores (\cref{sec:search}, \cref{alg:search}).}
\label{fig:overview}
\end{figure}

Text-to-image diffusion and flow models render most prompts well but still miss compositional ones: a wrong count, a missing object, a colour or position bound to the wrong object.
The standard test-time remedy, \mbox{best-of-$N$}, draws complete images and keeps the one a verifier prefers~\cite{ma2025inference,xie2025sana15}.
Every candidate is fully denoised before it is judged, even one doomed from the start, so the generator pays in full for images that search then discards.
Judging a candidate during denoising would stop that waste; what has stood in the way, we show, is not the image but the judge.

The outcome is visible long before denoising ends.
It is settled at the start, by the seed of a deterministic sampler or by the first steps of a stochastic one, and every denoising step yields the model's own clean-image estimate (\cref{eq:preview}); decoded, it is a blurry preview, but its layout is already in place: how many objects, which ones and where (\cref{fig:teaser}).
A judge that reads previews would let search stop doomed candidates early and spend compute where the outcome is still open.
Existing verifiers are not built for this: ImageReward~\cite{xu2023imagereward}, HPSv2.1~\cite{wu2023hpsv2}, PickScore~\cite{kirstain2023pickapic} and CLIP score~\cite{radford2021clip}, the ones in common use, are trained on finished images, and scoring intermediate estimates with them is unreliable~\cite{coca2025,zhang2025lpo,ttsnap2025}.
Search that scores clean-image estimates with them, as progressive seed pruning (PSP)~\cite{psp2026} and FK steering~\cite{singhal2025fk} do, inherits this blindness; the remedy so far is to train noise-aware scorers~\cite{ttsnap2025,drm2026}.

\begin{figure}[t]
\centering
\includegraphics[width=0.58\linewidth,trim=0.8 2.8 6.2 3.8,clip]{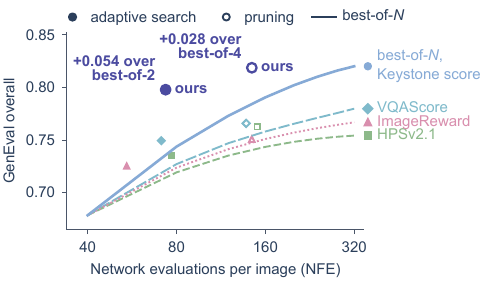}
\caption{\textbf{\method's searches lie above best-of-$N$ with every judge at the same cost} (SD3.5-Medium, official test prompts). Lines: best-of-$N$ on finished images with each judge; points: \method's adaptive search (filled) and pruning (open) with the same judge, each significantly more accurate with the \method score; numbers: paired gains over best-of-$N$ with the \method score (ours) at a higher cost, both significant.}
\label{fig:compute}
\end{figure}

\method (\cref{fig:overview}) changes the judge instead of training one.
It lets an off-the-shelf multimodal decision model~\cite{typesafe2026jev}, which answers questions about an image with a probability per answer, judge the previews directly.
Two things matter: the judge, and what it is asked. \method compiles each prompt into typed questions about its slots, such as whether there is an \emph{X}, how many, which colour and where, and scores an image by its weakest answer.
Asked this way, the decision model reads positions from step-8 previews better than VQAScore, ImageReward and HPSv2.1 read them from finished images, and counts better than ImageReward and HPSv2.1 do (\cref{tab:judges}).
A branch-and-read probe then finds the key steps at which the outcome is decided and from which it can be read, and two searches spend compute there: adaptive search draws further candidates only when the first preview falls short, and pruning keeps only the candidates whose previews score highest.
Search then stops finishing doomed candidates, and accuracy improves significantly at lower generator compute (\cref{fig:compute,tab:main}).

Our contributions are:
\begin{itemize}
\item \textbf{The outcome is visible long before denoising ends.} A branch-and-read probe shows that a stochastic sampler's first steps decide compositional outcomes, so search branches at the start (over seeds on a deterministic sampler), and that early previews already reveal them (\cref{sec:keysteps,sec:exp_keysteps}).
\item \textbf{A training-free judge that reads previews.} \method compiles the prompt into typed questions for an off-the-shelf decision model, which reads previews of counting, two-object and position prompts significantly better than every other judge (\cref{tab:judges}); the decision model carries most of the gain and the questions add on position (\cref{sec:exp_analysis}), and an open judge, Qwen3-VL-8B~\cite{bai2025qwen3vl}, answering their yes/no form also lifts both searches on SD3.5-Medium significantly above every verifier's best-of-$N$ at equal cost (\cref{sec:exp_judges}).
\item \textbf{Significant improvement at lower cost.} On GenEval with SD3.5-Medium, both searches are significantly more accurate than best-of-$N$ with every other verifier at a fraction of its generator compute (\cref{fig:compute} and \cref{sec:app_cat_ci}); without retuning, both remain significantly better than the strongest other verifier's best-of-$N$ on average across diffusion and flow models on GenEval, T2I-CompBench (\cref{sec:exp_general}) and GenAI-Bench (\cref{sec:app_gen}).
\item \textbf{A drop-in verifier for existing methods.} On GenEval with SD3.5-Medium, the \method score significantly improves best-of-$N$ at every $N$ over each other verifier, and PSP and FK steering (\cref{sec:exp_search}).
\end{itemize}

\section{Related Work}
\label{sec:related}

\paragraph{Test-time search for diffusion and flow models.}
Drawing several samples and keeping the one a verifier prefers already scales better than adding denoising steps~\cite{ma2025inference}, and with verifiers trained on GenEval-style data it reaches high GenEval scores at large sample counts~\cite{xie2025sana15,li2025reflectdit,zhuo2025reflectionflow}.
Multimodal agents can also improve the image over several rounds of visual reasoning~\cite{DBLP:journals/corr/abs-2602-06166}.
Search over the trajectory scores intermediate states to steer or resample: Feynman--Kac steering~\cite{singhal2025fk}, soft value-based decoding~\cite{li2024svdd}, dynamic search~\cite{li2025dsearch}, rollover budget forcing~\cite{kim2025rbf} and evolutionary search~\cite{he2025evosearch}.
Weighting sampled trajectories by their outcome is likewise the basis of path-integral planning in reinforcement learning~\cite{DBLP:journals/corr/abs-2011-06752}.
PSP~\cite{psp2026} starts many seeds, rates their clean-image estimates by ImageReward and keeps the best; at matched compute on SD3.5-Large, none of the ImageReward-guided methods it compares, PSP included, exceeds best-of-$N$ on GenEval.
TTSnap~\cite{ttsnap2025} and DRM~\cite{drm2026} train noise-aware scorers because clean-image reward models rank intermediate estimates inconsistently.
\method keeps the search simple and changes the judge: an off-the-shelf decision model, asked about the prompt's structure.

\paragraph{Test-time computation and adaptive inference.}
In language models, how much test-time computation can add~\cite{yang2026languagemodelsgaintesttime}, how to spend it efficiently~\cite{DBLP:journals/corr/abs-2604-02322} and whether improvements need more training at all~\cite{yang2026doesscalingreinforcementlearning,yang2026trainingfreeonpolicydistillation} have been studied directly.
Adaptive inference spends compute where an input needs it, by pruning tokens and channels in vision transformers~\cite{DBLP:journals/pami/LiTMFCMWZ26}, and by scheduling models and pruning tokens~\cite{DBLP:conf/iclr/LiMSJTFMXWZ26} or learning when to think~\cite{DBLP:journals/corr/abs-2605-29438} in vision--language--action models.
\method's adaptive search and pruning do the same for a diffusion sampler: the previews decide which candidates receive further denoising.

\paragraph{Process reward models.}
In language-model reasoning, a process reward model scores each intermediate step rather than only the final answer~\cite{uesato2022process}, and selecting among sampled solutions with it is more reliable than with an outcome reward model~\cite{lightman2023verify}.
Both are trained on step labels.
\method plays the same role for a flow sampler without any step-level training: the intermediate step is the model's own preview, and the prompt's slot questions supply what the step should already satisfy.
Reasoning process rewards likewise improve audio language models~\cite{DBLP:conf/iclr/FanRLPSBGLG26}, and audio reasoners can improve themselves recursively~\cite{fan2026audiorsirecursivelyselfimprovingaudio}; procedure-aware reinforcement learning trains tool-augmented language models~\cite{zheng2026procedureaware}, and an audio-native gym trains voice agents~\cite{DBLP:journals/corr/abs-2608-26432}.

\paragraph{Per-step signals in flow-model RL.}
Flow-matching models are fine-tuned online with reward weighting and Wasserstein regularization~\cite{DBLP:conf/iclr/FanSCCLL25}, and generative models with adaptive divergence-regularized policy optimization~\cite{DBLP:conf/nips/FanWCCL25}.
GRPO-style methods for flow models~\cite{liu2025flowgrpo,xue2025dancegrpo,li2025mixgrpo,zheng2025diffusionnft,xue2025awm,wang2025cps} give every step of a trajectory the same advantage.
Per-step credit comes from branching at intermediate states and finishing every branch~\cite{he2025tempflowgrpo,zhou2025g2rpo,li2025branchgrpo,ding2025treegrpo}, from intermediate feedback~\cite{DBLP:journals/corr/abs-2510-18072}, from scoring clean-image estimates with the training reward~\cite{stepwise2026,vgpo2025,densegrpo2026,tpgrpo2026}, or from reward models trained on noisy latents~\cite{liang2024spo,zhang2025lpo,drm2026}.
A judge that reads previews without training could supply such signals.

\paragraph{Judges for generated images.}
Preference models such as ImageReward~\cite{xu2023imagereward}, HPSv2.1~\cite{wu2023hpsv2} and PickScore~\cite{kirstain2023pickapic}, and CLIP similarity~\cite{radford2021clip}, return one scalar per image.
GenEval~\cite{ghosh2023geneval} checks compositional prompts with an object detector; RL methods that train on this detector as the reward reach near-perfect scores on it~\cite{liu2025flowgrpo,zheng2025diffusionnft,xue2025awm}.
Question-based metrics evaluate finished images by asking about the prompt: TIFA~\cite{hu2023tifa} and DSG~\cite{cho2024dsg} generate questions from the prompt and answer them with a visual question-answering model, and VQAScore~\cite{lin2024vqascore} asks whether the image shows the prompt.
\method's questions come from the prompt's slots in the same spirit, but serve as a process reward: they are asked about previews during sampling and decide where the sampler spends its compute.
For autoregressive image generation, PARM~\cite{guo2025parm} trains a step-wise reward model and finds that judging unfinished images naively does worse than judging finished ones.

\section{Method}
\label{sec:method}

\subsection{Previews and Judges}
\label{sec:previews}

A rectified-flow model generates an image by integrating a learned velocity field $v_\theta$ from noise ($\sigma=1$) to data ($\sigma=0$).
An Euler step from $\sigma_i$ to $\sigma_{i+1}$ is
\begin{equation}
x_{i+1} = x_i + (\sigma_{i+1}-\sigma_i)\, v_\theta(x_i, \sigma_i, c),
\end{equation}
and the same evaluation of $v_\theta$, under any solver, gives the model's own estimate of the clean image,
\begin{equation}
\hat x_0^{(i)} = x_i - \sigma_i\, v_\theta(x_i, \sigma_i, c).
\label{eq:preview}
\end{equation}
Decoding $\hat x_0^{(i)}$ with the autoencoder yields a \emph{preview} of the final image at no extra cost in network evaluations.
A diffusion model that predicts the noise gives the same estimate, $\hat x_0^{(i)} = x_i - \sigma_i\,\hat\epsilon_\theta(x_i, \sigma_i, c)$, which its Euler sampler computes anyway; we read SDXL's previews this way (\cref{sec:exp_general}).
Early previews are blurry, but the layout---how many objects and which ones---is often already in place.

\method's judge is an off-the-shelf decision model: given a prompt and an image, it answers a fixed set of questions about them in a single call, each with a probability distribution over the allowed answers.
We use Jev~\cite{typesafe2026jev} as released, without training, tuning or labels, and \cref{sec:exp_judges} shows that an open vision--language model can answer the yes/no form of the same questions.
Any judge that gives a probability per answer can take Jev's place; \method also fixes what the judge is asked, and when.
A question is typed by its answer: a probability that a statement holds, a choice among listed options, or a bounding box.

\subsection{Compiling Prompts into Typed Questions}
\label{sec:questions}

Compositional prompts have slots: which objects are required, how many of each, their colours, and where they are relative to each other.
GenEval~\cite{ghosh2023geneval} ships them with every prompt; elsewhere a language model writes them down (\cref{sec:exp_general}).
We compile the prompt into questions about its slots, each with a single, checkable answer:
\begin{itemize}
\item \textbf{Presence and count.} ``Is there at least one \emph{X}?''; for counting prompts, ``Are there exactly $n$ \emph{X}s?'' together with ``Are there $n{+}1$ or more?'', which must be false.
\item \textbf{Colour.} ``Is the \emph{X} \emph{c}?''; for colour attribution, a choice among the colour names GenEval uses, and an object's score is the smaller of its presence probability and the probability of its requested colour.
\item \textbf{Position.} ``Is the \emph{X} left of the \emph{Y}?'' and a box for each of the two objects; the boxes' relation is computed with GenEval's own rule (centre offset after a margin proportional to the object sizes), kept continuous for ranking, and averaged with the smallest yes/no answer.
\end{itemize}
A prompt is satisfied only if every slot holds, so a candidate's \emph{\method score} is the smallest of its slot scores.
The same questions are asked about previews and final images.

\subsection{Locating the Key Steps}
\label{sec:keysteps}

A branch-and-read probe asks two questions about a sampler: at which steps is the outcome decided, and from which step can it be read?
To isolate one step, we run a shared prefix to step $s$, let six siblings draw their own noise at step $s$ only, and continue all of them with shared randomness, so that any difference in how they end is caused by the choice at step $s$.
The \emph{decision curve} is the fraction of sibling groups whose official verdict differs between siblings, as a function of $s$.
The \emph{readability curve} is, for each read step $r$, the fraction of candidate pairs with different official verdicts on their final images that a judge, reading the previews at step $r$, orders correctly.
Together they give the key steps: the window where the first curve is high, and the read step after it, from which the second is already high.
On a deterministic sampler, which draws noise only at the start, the outcome is fixed by the seed, so search there runs over seeds and reads their previews at the read step.

\subsection{Test-Time Search at the Key Steps}
\label{sec:search}

Two searches use the \method score as gradient-free guidance: the previews' answers decide which trajectories receive further compute, without differentiating through the judge or changing the sampler.
We count cost as the number of network evaluations per generated image (NFE; a guided step counts once), as in the best-of-$N$ and search literature; the judge runs outside the generator and is reported separately in calls and dollars.
For a sampler with $T$ steps, best-of-$N$ costs $NT$.

\begin{algorithm}[t]
\caption{Search at the key steps. $\textsc{Run}(z, a, b)$ integrates candidate $z$ from step $a$ to step $b$; $\textsc{Ask}(z, r)$ is the \method score of the preview of $z$ at step $r$ (\cref{sec:questions}); $T$ is the number of sampler steps; comments give the cost in NFE.}
\label{alg:search}
\small
\begin{algorithmic}[1]
\Procedure{Adaptive}{$r, K, \tau$}
  \State $z_1 \gets$ new seed; \textsc{Run}$(z_1, 0, r)$
  \If{\textsc{Ask}$(z_1, r) \ge \tau$} \State \Return \textsc{Run}$(z_1, r, T)$ \Comment{$T$} \EndIf
  \For{$k = 2, \dots, K$} \State $z_k \gets$ new seed; \textsc{Run}$(z_k, 0, r)$ \EndFor
  \State \Return \textsc{Run}$(\arg\max_k \textsc{Ask}(z_k, r), r, T)$ \Comment{$Kr{+}T{-}r$}
\EndProcedure
\Procedure{Prune}{$K_0, (r_1, k_1), \dots, (r_m, k_m)$}
  \State $S \gets K_0$ new seeds; $r_0 \gets 0$
  \For{$i = 1, \dots, m$}
    \State \textsc{Run}$(z, r_{i-1}, r_i)$ for $z \in S$
    \State $S \gets$ the $k_i$ members of $S$ with the highest \textsc{Ask}$(z, r_i)$
  \EndFor
  \State \textsc{Run}$(z, r_m, T)$ for $z \in S$
  \State \Return the member of $S$ whose final image scores highest
\EndProcedure
\end{algorithmic}
\end{algorithm}

\paragraph{Adaptive search.}
Run one candidate to the read step $r$ and ask about its preview.
If its score reaches a threshold $\tau$, finish it, for a total of $T$.
Otherwise run $K{-}1$ further candidates to step $r$, finish the one with the highest score, and pay $Kr + (T-r)$; the budget thus goes to the images whose first preview falls short.

\paragraph{Pruning.}
Start $K_0$ candidates, read them at step $r_1$ and keep the best $k_1$, read again at $r_2$ and keep $k_2$, and so on; survivors are finished and, if more than one remains, compared on their final images.
Its structure is the progressive seed pruning of PSP~\cite{psp2026}; \method replaces PSP's reward model with its own score and reads at the key steps. \cref{alg:search} states both procedures.

\section{Experiments}
\label{sec:exp}

\begin{figure}[t]
\centering
\includegraphics[width=\linewidth,trim=0.0 0.0 0.0 2.0,clip]{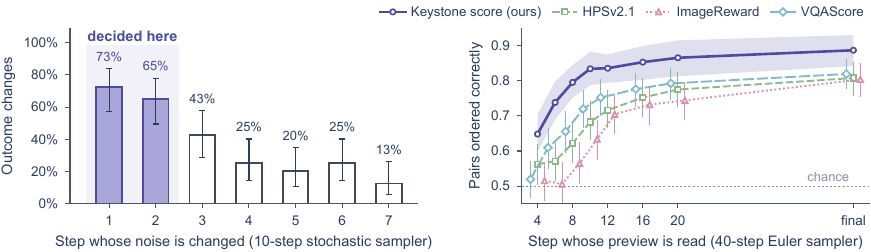}
\caption{\textbf{Compositional outcomes are decided in the first steps and can be read well before the end} (SD3.5-Medium, development prompts).
Left: fraction of the 40 sibling groups whose GenEval verdicts differ when only the noise of one step of a 10-step stochastic sampler (Flow-GRPO's~\cite{liu2025flowgrpo}) changes.
Right: fraction of candidate pairs with different verdicts that each judge orders correctly from the preview at each step of the 40-step schedule, counting and two-object prompts pooled.
Left, error bars: 95\% Wilson intervals; right, band and bars: 95\% bootstrap intervals over prompts.}
\label{fig:keysteps}
\end{figure}

\paragraph{Experimental setup.}
We use SD3.5-Medium~\cite{esser2024sd3} with Flow-GRPO's evaluation setting~\cite{liu2025flowgrpo} (40 Euler steps) and evaluate on GenEval~\cite{ghosh2023geneval} with its official detector and code, unmodified (full protocol: \cref{sec:app_details}).
Every setting---read steps, thresholds, numbers of candidates, pruning schedules, and the way each slot is asked---is chosen on development prompts from GenEval's own prompt generator with a different seed, disjoint from the test prompts, and applied to them unchanged.
A gain is significant when its 95\% paired bootstrap interval over prompts lies above zero.
Judges compared on one model read the same candidates: on SD3.5-Medium, the \method score, VQAScore~\cite{lin2024vqascore}, ImageReward~\cite{xu2023imagereward} and HPSv2.1~\cite{wu2023hpsv2} read previews and final images, and PickScore~\cite{kirstain2023pickapic} and CLIP score~\cite{radford2021clip} final images only.

\subsection{Key-Step Analysis}
\label{sec:exp_keysteps}

\begin{figure}[t]
\centering
\includegraphics[width=0.58\linewidth,trim=0.0 2.5 3.0 0.0,clip]{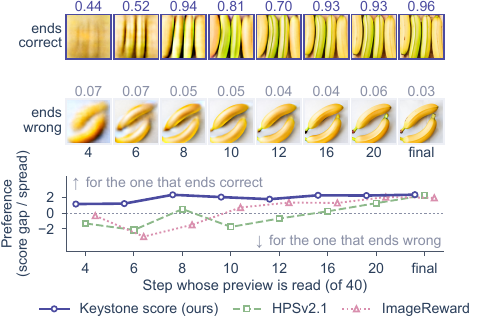}
\caption{\textbf{Case study: \method ranks the candidate that ends correct above the other at every step} (the first and last step-8 candidates of \cref{fig:teaser}). Top: previews and final image with their \method score. Bottom: each judge's preference between the two, the difference of its two scores over the spread of its scores at that step.}
\label{fig:film}
\end{figure}

\textbf{Compositional outcomes are decided in the first steps and can be read long before the end} (\cref{fig:keysteps}).
Changing the noise of a single step of the stochastic schedule alters the verdict on the finished image far more often at one of the first two steps than at steps 4 to 7 (left).
Pooled over counting and two-object prompts, the \method score orders candidates by their previews significantly better than every other judge reading the same previews, at every step (right; per category at step 8: \cref{tab:judges}); on the test prompts its previews from step 12 on also beat every other judge's finished images (\cref{sec:app_nonsig}).
Search should therefore branch at the start---over seeds on the deterministic sampler we search---and read shortly after; the development prompts select \method's read steps 8 and 16.

\paragraph{Case study.}
In \cref{fig:film}, the \method score ranks the first step-8 candidate of \cref{fig:teaser}, which ends correct, above the last one at every recorded step; ImageReward prefers the wrong one up to step 8, and HPSv2.1 switches back and forth.

\subsection{Preview Readability Across Judges}
\label{sec:exp_judges}

\begin{figure}[t]
\centering
\includegraphics[width=0.58\linewidth,trim=4.8 4.5 1.8 4.8,clip]{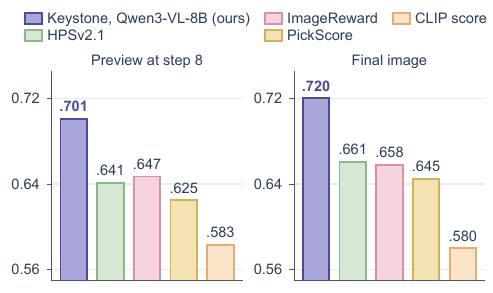}
\caption{\textbf{On Sana-0.6B, an open judge answering \method's questions reads previews best} (GenEval counting and two-object; six candidates per prompt that branch after step 2 of 20). Pairs of candidates ordered correctly from the step-8 preview (left) and from the final image (right), with \method's questions answered by Qwen3-VL-8B. Every judge and reading: \cref{sec:app_sana_judges}.}
\label{fig:sana}
\end{figure}

\begin{figure}[t]
\centering
\includegraphics[width=\linewidth]{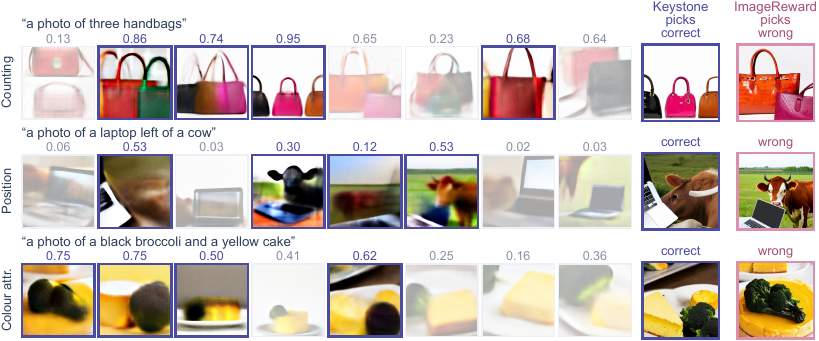}
\caption{\textbf{\method's pruning ends correct where ImageReward's ends wrong} (SD3.5-Medium, official test prompts). Per prompt, an image's eight step-8 candidates with \method scores (framed: the four kept) and the image each pruning ends on (correct/wrong: verdicts of GenEval's official evaluator, as in \cref{fig:teaser}). Examples follow a fixed rule (\cref{sec:app_qual_more}); this happens for \qualWinAll{} images in these categories.}
\label{fig:qual}
\end{figure}

\textbf{\method reads counts, objects and positions from previews best, while on identical candidates the other judges fall to near chance for several categories} (\cref{tab:judges}); HPSv2.1 does not read relative position even from finished images.
Reading the same previews, the \method score is significantly better than every other judge on counting, two-object and position prompts, and than both reward models on colours; at step 8 it reads positions better than every other judge of \cref{tab:judges} reads finished images, and counts better than both reward models do (intervals: \cref{sec:app_judge_ci}).
On the official test prompts, whose candidates were generated only after these settings were fixed, it again orders step-8 previews better than both reward models in every category, significantly (\cref{sec:app_judge_ci}).

\begin{table}[t]
\centering
\small
\setlength{\tabcolsep}{4pt}
\begin{tabular}{lcccccccc}
\toprule
 & \multicolumn{5}{c}{Preview at step 8} & \multicolumn{3}{c}{Final image} \\
\cmidrule(lr){2-6}\cmidrule(lr){7-9}
Judge & Two obj. & Count. & Colours & Pos. & Col. attr. & Count. & Colours & Pos. \\
\midrule
\ours \method score (ours) & \textbf{0.73} & \textbf{0.86} & \textbf{0.62} & \textbf{0.87} & \textbf{0.57} & \textbf{0.93} & \textbf{0.68} & \textbf{0.97} \\
VQAScore & 0.57 & 0.74 & 0.56 & 0.57 & 0.51 & 0.82 & 0.64 & 0.79 \\
HPSv2.1 & 0.58 & 0.66 & 0.50 & 0.52 & 0.51 & 0.77 & 0.64 & 0.52 \\
ImageReward & 0.61 & 0.52 & 0.52 & 0.57 & 0.48 & 0.72 & 0.63 & 0.66 \\
\bottomrule
\end{tabular}

\caption{\textbf{\method reads counts, objects and positions from step-8 previews best} (SD3.5-Medium, development prompts). Fraction of candidate pairs with different official verdicts that each judge orders correctly, from the step-8 preview and, for three categories, the final image; bold: best (step-8 intervals and test prompts: \cref{sec:app_judge_ci}).}
\label{tab:judges}
\end{table}

\paragraph{An open judge asked \method's questions also reads previews best.}
Any judge that gives a probability per answer can take Jev's place (\cref{sec:previews}).
On the smaller flow model \mbox{Sana-0.6B}~\cite{xie2024sana}, the open vision--language model Qwen3-VL-8B answering \method's questions reads previews and finished images significantly better than HPSv2.1, ImageReward, PickScore and CLIP score (\cref{fig:sana}); as the judge of both searches on SD3.5-Medium, it keeps them significantly above best-of-$N$ with every verifier at the same cost (\cref{sec:app_open_judge}).

\subsection{Main Results: Test-Time Search on GenEval}
\label{sec:exp_search}

\textbf{Both searches are more accurate than best-of-$N$ with every verifier at lower generator compute} (\cref{tab:main,fig:compute}).
Reading previews, they stop candidates whose layout is already wrong instead of finishing them, and spend the saved compute where the outcome is still open; \cref{tab:main} sets them against the strongest verifiers, VQAScore and the \method score itself, and against PSP and FK steering.
Every form that uses the \method score---best-of-$N$ with it, adaptive search and pruning---is ours; the other verifiers drive the same procedures for comparison (\cref{fig:qual}).
Every gain stated below is significant by the paired bootstrap over prompts defined above.

\begin{table}[t]
\centering
\small
\setlength{\tabcolsep}{6pt}
\begin{tabular}{lcccc}
\toprule
Method & NFE & Count. & Position & Overall \\
\midrule
One sample & 40 & 0.56 & 0.26 & 0.68 \\
\midrule
Best-of-2, VQAScore & 80 & 0.66 & 0.32 & 0.73 \\
\ours Best-of-2, \method score (ours) & 80 & 0.70 & 0.37 & 0.74 \\
Adaptive search$^\ddagger$, VQAScore & 71 & 0.71 & 0.39 & 0.75 \\
\ours \method, adaptive (ours) & 73 & 0.79 & 0.53 & \textbf{0.80} \\
\midrule
Best-of-4, VQAScore & 160 & 0.72 & 0.37 & 0.76 \\
\ours Best-of-4, \method score (ours) & 160 & 0.78 & 0.48 & 0.79 \\
PSP~\cite{psp2026}, ImageReward & 152 & 0.70 & 0.32 & 0.75 \\
FK steering~\cite{singhal2025fk}, ImageReward & 160 & 0.67 & 0.31 & 0.71 \\
Pruning$^\ddagger$, VQAScore & 144 & 0.78 & 0.35 & 0.77 \\
\ours \method, pruning (ours) & 144 & 0.83 & 0.57 & \textbf{0.82} \\
\bottomrule
\end{tabular}

\caption{\textbf{\method's searches are the most accurate at each budget} (SD3.5-Medium, official test prompts, four images per prompt, each a search over eight candidates). Each block: methods that cost no more than its best-of-$N$, against the strongest verifiers (every verifier: \cref{sec:app_main_full}); bold: best overall; $^\ddagger$: \method's search with another judge. Each row with the \method score is significantly above its procedure with every other verifier. Pruning keeps four of eight at step 8 and two at step 16; PSP~\cite{psp2026} and FK steering~\cite{singhal2025fk} use ImageReward. Gains are computed before rounding.}
\label{tab:main}
\end{table}

\begin{figure}[tbp]
\centering
\begin{minipage}[b]{0.48\linewidth}\centering
\includegraphics[width=\linewidth,trim=0.8 2.5 1.8 5.5,clip]{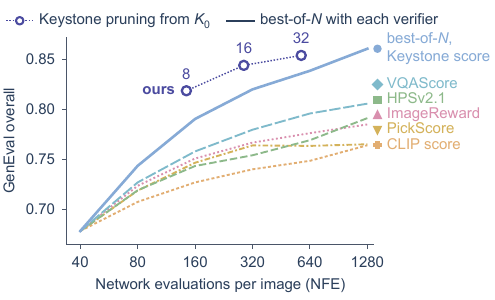}
\end{minipage}\hfill
\begin{minipage}[b]{0.48\linewidth}\centering
\includegraphics[width=\linewidth,trim=3.2 7.0 1.0 5.2,clip]{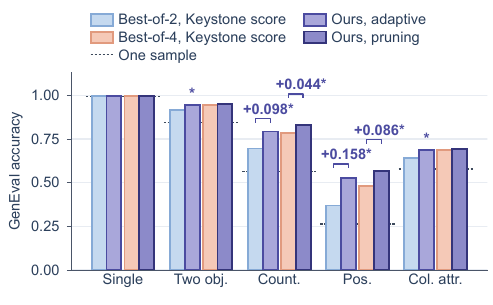}
\end{minipage}
\par\vspace{0.4em}
\begin{minipage}[t]{0.48\linewidth}
\caption{\textbf{\method's pruning from $K_0$ candidates beats every other verifier's best-of-$K_0$} (SD3.5-Medium, official test prompts). Lines: best-of-$N$ with each verifier, $N$ up to 32; circles: \method's pruning from $K_0=8$, 16 and 32 candidates.}
\label{fig:scaling}
\end{minipage}\hfill
\begin{minipage}[t]{0.48\linewidth}
\caption{\textbf{The gain is largest on counting and position} (SD3.5-Medium, official test prompts). Accuracy per category; best-of-$N$ uses the \method score (ours) on finished images; dashed: one sample; brackets: gain over the best-of-$N$ beside it; $\ast$: paired 95\% interval above zero.}
\label{fig:gaincat}
\end{minipage}
\end{figure}

\paragraph{\method's searches are the most accurate at each budget.}
Within each budget of \cref{tab:main}, adaptive search and pruning are above best-of-$N$ with every judge and above the same search driven by any other judge, and pruning is above every other verifier's best-of-8 at under half its cost (settings and every comparison: \cref{sec:app_settings,sec:app_cat_ci}).
Started from more candidates, pruning likewise beats every other verifier's best-of-$N$ with as many candidates at a fraction of its cost (\cref{fig:scaling}).
PSP and FK steering, rerun with ImageReward, fall below it at a higher cost, and unlike the published test-time methods that PSP compares, it exceeds best-of-$N$ at lower compute (\cref{sec:app_published}).

\begin{figure}[tbp]
\centering
\begin{minipage}[b]{0.48\linewidth}\centering
\includegraphics[width=\linewidth,trim=8.5 1.8 2.2 4.8,clip]{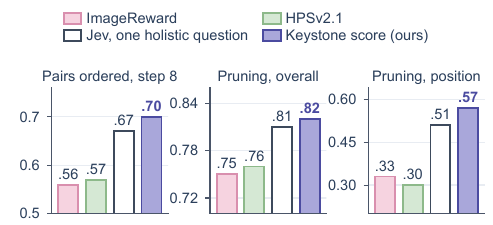}
\end{minipage}\hfill
\begin{minipage}[b]{0.48\linewidth}\centering
\includegraphics[width=\linewidth,trim=2.5 2.0 5.8 2.2,clip]{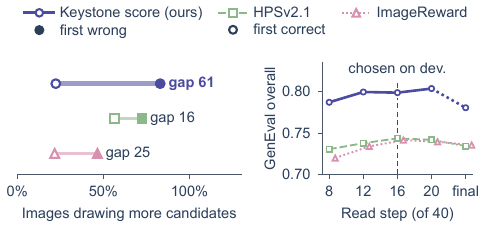}
\end{minipage}
\par\vspace{0.4em}
\begin{minipage}[t]{0.48\linewidth}
\caption{\textbf{A decision model carries most of the gain; the questions add on position} (SD3.5-Medium, official test prompts). The same judge asked one holistic question or \method's questions, beside both reward models, on the same candidates; pruning uses the schedule of \cref{tab:main}.}
\label{fig:ablation}
\end{minipage}\hfill
\begin{minipage}[t]{0.48\linewidth}
\caption{\textbf{\method spends extra candidates where the first one fails} (SD3.5-Medium, official test prompts). Left: share of images that draw more candidates when the first ends wrong or correct. Right: best accuracy within best-of-2's cost per read step, the envelope over the development grid of $K$ and $\tau$, formed alike for every judge.}
\label{fig:adaptive}
\end{minipage}
\end{figure}

\paragraph{The \method score also improves the existing methods it is plugged into.}
At the same cost and on the same candidates, best-of-$N$ is significantly more accurate with it than with each other verifier at every $N$ from 2 to 8, and PSP and FK steering both gain significantly when it replaces their ImageReward (\cref{sec:app_dropin}); only the verifier changes, so the gain comes from the judge.

\paragraph{Reading previews adds most on counting and position.}
Best-of-$N$ with the \method score reads the same questions from finished images, so comparing each search with it isolates what reading previews adds (\cref{fig:gaincat}): both searches gain most on the slots the \method score reads best from previews (\cref{tab:judges}); adaptive search also gains on colour attribution and two-object prompts (\cref{sec:app_cat_ci}).
In the examples of \cref{fig:qual}, pruning removes candidates whose layout is already wrong at the read step, and comparing the survivors' final images settles the rest.
Driven by another judge, the same pruning schedule is significantly less accurate; the judge runs outside the generator and adds no generator compute: adaptive search calls it a few times per image and pruning 14 times per image (\cref{sec:app_sameproc,sec:app_cost_practice}).

\subsection{Ablation Study and Analysis}
\label{sec:exp_analysis}

\paragraph{A decision model carries most of the gain; the compiled questions add on position.}
To separate the judge from the questions, we ask Jev one holistic question per preview---does the image show everything the prompt asks for---in the spirit of VQAScore~\cite{lin2024vqascore}, with images, read steps and pruning schedule fixed (\cref{fig:ablation}).
With the one question, Jev already orders step-8 preview pairs better than either reward model and prunes to higher accuracy, both significantly (\cref{sec:app_ablcat}).
The compiled questions add to this where one question is too coarse: they order these pairs better than the holistic question and, in pruning and in picking final images, raise accuracy on position, where they also ask for boxes and apply GenEval's rule, all significantly (\cref{sec:app_ablcat}).

\paragraph{\method spends extra candidates where the first one fails.}
In adaptive search, the judge decides where generator compute goes (\cref{fig:adaptive}, left): \method draws more candidates for most images whose first candidate will fail and for far fewer of those whose first will succeed, a gap significantly wider than either reward model's (\cref{sec:app_alloc}), so the budget goes to the images whose first preview falls short.
The read step is not a sensitive choice (\cref{fig:adaptive}, right): every read step is above reading only final images at the same budget, the curve is nearly level around the step chosen on the development prompts, and even at the earliest read step, \method's best setting lies above either reward model's best at any step.
Finishing candidates one at a time until a final image passes a threshold is likewise significantly below adaptive search at its cost, so within this budget the saving comes from reading previews (\cref{sec:app_readfinal}).

\begin{figure}[tbp]
\centering
\begin{minipage}[b]{0.48\linewidth}\centering
\includegraphics[width=\linewidth,trim=2.0 2.2 1.2 5.2,clip]{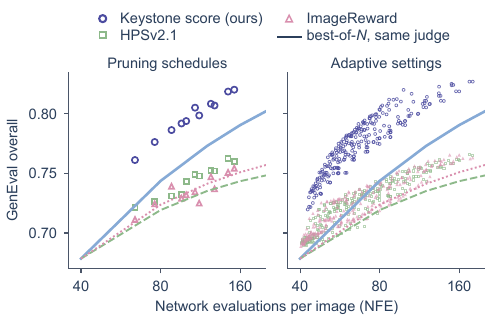}
\end{minipage}\hfill
\begin{minipage}[b]{0.48\linewidth}\centering
\includegraphics[width=\linewidth,trim=0.0 6.2 2.2 5.2,clip]{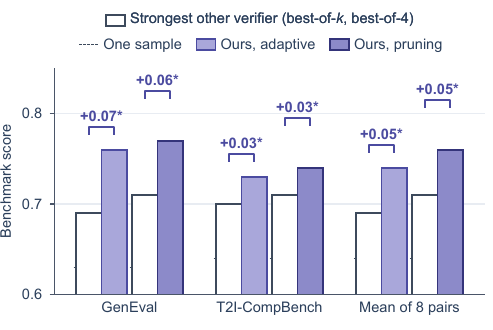}
\end{minipage}
\par\vspace{0.4em}
\begin{minipage}[t]{0.48\linewidth}
\caption{\textbf{Every setting driven by the \method score beats best-of-$N$} (SD3.5-Medium, official test prompts). Points: every pruning schedule and adaptive setting, driven by each judge; lines: best-of-$N$ with the same judge.}
\label{fig:robust}
\end{minipage}\hfill
\begin{minipage}[t]{0.48\linewidth}
\caption{\textbf{Without retuning, the gain holds on average over four base models and two benchmarks} (mean over SD3.5-Medium, FLUX.1-dev, SDXL and Sana-0.6B). Best-of-$k$ and best-of-4: the strongest other verifier at no lower cost than adaptive search and pruning; brackets: our gain; $\ast$: pooled paired 95\% interval above zero. GenAI-Bench: \cref{sec:app_gen}.}
\label{fig:general}
\end{minipage}
\end{figure}

\paragraph{The gain does not rest on one tuned setting.}
Driven by the \method score, every pruning schedule fixed in advance and every adaptive setting of the development grid significantly beats best-of-$N$ with it at equal cost, and every schedule significantly beats the same schedule driven by either reward model (\cref{fig:robust}).

\subsection{Other Models and Benchmarks}
\label{sec:exp_general}

\textbf{Without retuning, the gain holds on average with three more base models and two more benchmarks} (\cref{fig:general}): the large flow transformer FLUX.1-dev~\cite{flux2024}, the $\epsilon$-prediction diffusion U-Net SDXL~\cite{podell2023sdxl} and Sana-0.6B, and the benchmarks T2I-CompBench~\cite{huang2023t2icompbench} and GenAI-Bench~\cite{li2024genaibench}.
A language model writes their slots once, before generation, and their own metrics, not GenEval's rule, score the images (\cref{sec:app_bench,sec:app_questions}).
Averaged over model--benchmark pairs, both searches beat the strongest other verifier's best-of-$N$ significantly on all three benchmarks (\cref{fig:general}; GenAI-Bench: \cref{sec:app_gen}), so the gain is more than agreement with GenEval's evaluator.
The key steps carry over: on every base model, the \method score reads previews significantly better than the reward models do at every read step, including the read steps fixed on SD3.5-Medium (\cref{sec:app_readgen}). Against HPSv3~\cite{ma2025hpsv3}, a newer preference model, the \method score is significantly the better verifier of best-of-$N$ at every $N$ we test, and both searches exceed HPSv3's best-of-8 at under half its cost (\cref{sec:app_hpsv3}).

\section{Conclusion}
\label{sec:conclusion}

The outcome of a compositional prompt is visible long before denoising ends: it is settled at the start, and the model's own previews show it to a judge that asks about the prompt's structure.
\method is that judge, and searching at the key steps it spends generator compute only where the outcome is still open: on GenEval with SD3.5-Medium it significantly outperforms best-of-8 with every other verifier at under half its generator compute, and as a drop-in verifier its score significantly improves best-of-$N$, progressive seed pruning and FK steering; the searches' improvement holds without retuning on average across four diffusion and flow models and three benchmarks.
What held back early judgment was thus the judge, not the image: a released judge reads the sampler's own previews without any training on noisy latents, so any diffusion or flow sampler with a clean-image estimate can be searched at its key steps, and on GenEval with SD3.5-Medium an open judge still lifts both searches above every verifier's best-of-$N$ at equal cost.
The same judgment of previews could supply the per-step signal that reinforcement learning of flow models still lacks.

\pdfbookmark[1]{References}{references}
\bibliographystyle{unsrtcase}
\bibliography{main}

\clearpage
\appendix
\crefalias{section}{appendix}\crefalias{subsection}{subappendix}\crefalias{subsubsection}{subsubappendix}

\clearpage
\section{Analysis and Discussion}
\label{sec:app_discussion}

\subsection{Why the Outcome Can Be Read Early}
\label{sec:app_why}

A compositional prompt is judged by its layout: which objects appear, how many of each, which colour each has, and where they stand relative to each other.
In a flow or diffusion sampler, the clean-image estimate of \cref{eq:preview} is the model's running answer to what the finished image will be, and the layout is the first thing that answer commits to.
The decision curve of \cref{fig:keysteps} locates the commitment: on the stochastic sampler, a new draw at one of the first steps changes the verdict far more often than one at a later step, and on the deterministic sampler the seed alone fixes it.
What an early preview still lacks---texture, edges, fine detail---is what no compositional slot asks about.

This explains both sides of \cref{tab:judges}.
Reward models are trained to compare finished images, so their scores respond to the very cues that previews lack, and on several slots they rank previews near chance.
A decision model asked a slot question---is there a dog, are there exactly three, is the cup left of the plate---needs only the layout, which the preview already shows.
The \method score therefore reads counts and positions from step-8 previews better than reward models read them from finished images, and on every base model it reads previews better than the reward models read the same previews (\cref{sec:app_readgen}).
Intermediate states of other large models can likewise be read: the middle layers of an audio language model hold a workspace that can be verbalized~\cite{DBLP:journals/corr/abs-2608-24958}.

\clearpage
\subsection{What the Judge and the Questions Each Contribute}
\label{sec:app_contrib}

\method changes two things at once: who judges and what is asked.
The ablation of \cref{fig:ablation} separates them.
A decision model asked a single question about the whole prompt already orders preview pairs better than both reward models and prunes to higher accuracy; the judge supplies reliable answers on blurry inputs.
The compiled questions add to this where one question is too coarse: they make each answer specific to one slot and, for position, ask for boxes and apply GenEval's own relation rule to them (\cref{sec:app_ablcat}).
The two parts are complementary, and both come without training.

\clearpage
\subsection{Relation to Search Methods and Process Rewards}
\label{sec:app_relation}

Test-time search for diffusion and flow models has mostly varied the search---how trajectories are branched, resampled, steered or pruned~\cite{singhal2025fk,li2024svdd,li2025dsearch,kim2025rbf,he2025evosearch,psp2026}---while keeping a verifier trained on finished images.
\method varies the judge and keeps the search simple, and the experiments show where the gain comes from.
The same pruning schedule driven by a reward model gains far less (\cref{tab:app_main_full}); PSP's own schedule driven by the \method score needs more compute to reach the accuracy of \method's pruning; and every search that reads a score can take the \method score in place of its verifier: best-of-$N$, PSP and FK steering all improve significantly when it does (\cref{sec:app_dropin}).
\method is therefore complementary to work on search algorithms: a better schedule and a better judge compose.
Deciding which candidates receive further compute is also a selection problem, akin to the behavior selection and data-distribution optimization that make deep reinforcement learning sample-efficient on Atari~\cite{DBLP:conf/iclr/FanZLH0ZWX23,DBLP:conf/icml/FanX22}.

Process reward models in language-model reasoning score intermediate steps and are trained on step labels~\cite{uesato2022process,lightman2023verify}.
\method obtains step-level judgments without step labels: the sampler's intermediate state is decoded into an image the judge can read, and the prompt states what that image must already satisfy.

\clearpage
\subsection{Cost in Practice}
\label{sec:app_cost_practice}

The judge runs beside the generator, not inside it: it adds no network evaluations, needs no gradients, and leaves the sampler unchanged, so every NFE reported in the paper counts the generator alone.
Its price is about \adaptCalls{} calls per image for adaptive search and \pruneCalls{} for pruning, a fraction of a cent (\cref{sec:app_cost}); the calls of one read run in parallel, so a search waits for at most two rounds of calls (adaptive search) or three (pruning).
Generator evaluations are what best-of-$N$ and every search spend; \method spends fewer of them and gains accuracy.

\clearpage
\subsection{Broader Use}
\label{sec:app_broader}

Nothing in \method is specific to one model or benchmark.
Any sampler that computes a clean-image estimate can be searched, as the four base models of \cref{sec:app_gen} show on average, from flow transformers to an $\epsilon$-prediction U-Net; prompts without annotated slots are compiled by a language model once, before generation (\cref{sec:app_questions}); and on SD3.5-Medium an open judge, Qwen3-VL-8B, answering only the yes/no questions, also drives both searches above every verifier's best-of-$N$ at the same cost (\cref{sec:app_open_judge}).
Flow models now also generate discrete data~\cite{DBLP:journals/corr/abs-2504-10283} and proteins of variable length~\cite{DBLP:journals/corr/abs-2607-09039}, and online reinforcement learning improves protein generators~\cite{DBLP:journals/tmlr/WangFGNJL26}; where such a sampler exposes its estimate of the final sample and the target can be checked by questions, the same reading could guide search.
The same per-step judgment could also serve training: GRPO-style methods for flow models assign every step of a trajectory the same advantage, and a judge that reads previews without training could supply the per-step signal they lack (\cref{sec:related}).

\clearpage
\section{Pseudocode of the Score and the Probe}
\label{sec:app_alg}

\Cref{alg:app_score} computes the \method score of one image, a preview or a final image (\cref{sec:questions}, \cref{sec:app_questions}); \cref{alg:app_probe} computes the two curves of the branch-and-read probe (\cref{sec:keysteps}); the two searches that use them are \cref{alg:search} of the main text.

\begin{algorithm}[!htb]
\caption{The \method score of an image $I$ for a prompt $p$ with slots: objects $o_i$ (class, count $n_i$, optional colour $c_i$, optional relation to another object), exclusions (a class with count $n{+}1$, for counting prompts) and GenEval's tag. $P(q)$ is the judge's probability that the yes/no question $q$ holds; $\sigma$ is the logistic function.}
\label{alg:app_score}
\small
\begin{algorithmic}[1]
\Procedure{Score}{$p, I$}
  \State $Q^{+}, Q^{-} \gets \emptyset$ \Comment{questions that must hold / must not}
  \For{each object $o_i$}
    \State add ``exactly $n_i$ $o_i$?'' to $Q^{+}$ if $p$ is a counting prompt, else ``at least $n_i$ $o_i$?''
    \State \textbf{if} $c_i$: add ``is $o_i$ $c_i$?'' to $Q^{+}$ and ask a colour choice for $o_i$
    \State \textbf{if} $o_i$ has relation $\rho$ to $o_j$: add ``is $o_i$ $\rho$ $o_j$?'' to $Q^{+}$ and ask boxes $b_i, b_j$
  \EndFor
  \State add the prompt's other adjectives and relations, if any, to $Q^{+}$ \Comment{other benchmarks}
  \State add ``$n{+}1$ or more of the class?'' to $Q^{-}$ for each exclusion
  \State ask every question about $I$ and $p$ in one call to the judge
  \State $y \gets \min\big(\{P(q)\}_{q\in Q^{+}} \cup \{1-P(q)\}_{q\in Q^{-}}\big)$ \Comment{every slot must hold}
  \If{tag $=$ position}
    \State $m \gets$ GenEval's relation margin of $(b_i, b_j)$ for $\rho$
    \State \Return $\tfrac12\big(y + \min_i P(\text{presence of } o_i)\,\sigma(10(m-\tfrac12))\big)$
  \ElsIf{tag $=$ colour attribution}
    \State \Return $\min_i \min\big(P(\text{presence of } o_i),\ P(\text{colour } c_i \text{ chosen for } o_i)\big)$
  \EndIf
  \State \Return $y$
\EndProcedure
\end{algorithmic}
\end{algorithm}

\begin{algorithm}[!htb]
\caption{The branch-and-read probe on development prompts. $v(\cdot)\in\{0,1\}$ is GenEval's official verdict on a final image; $\textsc{Score}_J$ is a judge's score.}
\label{alg:app_probe}
\small
\begin{algorithmic}[1]
\Procedure{DecisionCurve}{$s$} \Comment{stochastic sampler, probe step $s$}
  \For{each prompt}
    \State run a shared prefix to step $s$
    \State six siblings draw their own noise at step $s$ only and continue with shared noise to the end
    \State the group \emph{differs} if the siblings' verdicts $v$ are not all equal
  \EndFor
  \State \Return the fraction of groups that differ
\EndProcedure
\Procedure{ReadabilityCurve}{$r, J$} \Comment{read step $r$, judge $J$}
  \For{each pair $(a, b)$ of candidates of one image with $v(a)=1$, $v(b)=0$}
    \State credit $1$ if $\textsc{Score}_J(\text{preview}_r(a)) > \textsc{Score}_J(\text{preview}_r(b))$, $\tfrac12$ if equal, else $0$
  \EndFor
  \State \Return the mean credit
\EndProcedure
\end{algorithmic}
\end{algorithm}

The key steps are the window where the decision curve is high and the read step after it, where the readability curve is already high; the read steps themselves (8 and 16 of 40) and every other search setting are then chosen on the development prompts by search accuracy (\cref{sec:app_settings}).

\clearpage
\section{Implementation and Evaluation Details}
\label{sec:app_details}
\subsection{Base Models and Samplers}
\label{sec:app_protocol}

\paragraph{Samplers.}
SD3.5-Medium is sampled with 40 Euler steps of the rectified-flow ODE, classifier-free guidance 4.5, $512{\times}512$ pixels and T5 prompts truncated to 128 tokens, Flow-GRPO's evaluation setting~\cite{liu2025flowgrpo}.
We write the sampler out so that the model's clean-image estimate can be decoded at any step; given the same starting noise, its final images match those of the diffusers pipeline up to bfloat16 rounding.
SD3.5-Medium candidates are independent seeds; candidate $m$ of test prompt $n$ uses seed $10^6+1000n+m$, so every method in a comparison except FK steering (\cref{sec:app_verifiers}) reads the same candidates.
For \cref{fig:sana}, Sana-0.6B is sampled at $512{\times}512$ with 20 steps of a marginal-preserving rectified-flow SDE (guidance 4.5): for each prompt and seed, six candidates share the first two steps, each draws its own noise at the third, and all then continue with one shared noise sequence, so they differ only by that draw; previews are read at step 8 of 20.

\paragraph{Models.}
Besides SD3.5-Medium we search with three base models of different families, all sampled with 40 steps so that the read steps fixed on SD3.5-Medium (8 and 16 for pruning, the adaptive read step) mean the same fraction of the trajectory:
FLUX.1-dev~\cite{flux2024}, a 12B flow transformer ($512{\times}512$, distilled guidance 3.5, T5 prompts of up to 512 tokens);
SDXL~\cite{podell2023sdxl}, an $\epsilon$-prediction U-Net ($1024{\times}1024$, Euler, guidance 5.0);
and Sana-0.6B~\cite{xie2024sana}, a flow model with a linear-attention transformer ($512{\times}512$, its flow DPM-Solver, guidance 4.5; the 20-step setting of \cref{fig:sana} is not used here).
A preview is the model's own clean-image estimate: \cref{eq:preview} of the main text for the flow models and the scheduler's predicted original sample for SDXL.
Judges read SDXL's previews and final images, and VQAScore (also as GenAI-Bench's metric) its final images, at $512{\times}512$; GenEval's and T2I-CompBench's evaluators score its $1024{\times}1024$ final images.
Each prompt has eight candidates (seed $10^6+1000n+m$ for candidate $m$ of prompt $n$), searched once to give one image.

\clearpage
\subsection{Benchmarks, Prompts and Evaluator}
\label{sec:app_bench}

\paragraph{Development and test prompts.}
Development prompts are produced by GenEval's own prompt generator with a different random seed and share no prompt with the 553 test prompts; they cover five of the six categories, since the single-object prompts, one per object class, are the same under any seed.
There are 432 development prompts (two objects 99, counting 54, colours 79, position 100, colour attribution 100), each with eight candidates.
Each test prompt has 32 candidates, four images of eight candidates each; every image is an independent search over its own eight candidates.
Previews are decoded at steps 8, 12, 16 and 20 of 40 on the test prompts, and additionally at steps 4, 6 and 10 on the development prompts.

\paragraph{Benchmarks.}
GenEval~\cite{ghosh2023geneval}: the 553 test prompts and the official detector.
T2I-CompBench~\cite{huang2023t2icompbench}: 60 validation prompts of each of colour, shape, texture, spatial and numeracy, scored by the benchmark's own metric for the category (BLIP-VQA for the three attribute categories, UniDet for spatial and numeracy); spatial prompts are drawn among those whose objects UniDet can name, since every image of the others scores zero.
GenAI-Bench~\cite{li2024genaibench}: 150 basic and 150 advanced prompts drawn with a fixed seed from its 527 image prompts, scored by its metric, VQAScore with CLIP-FlanT5-XXL on the final image; there, VQAScore is the benchmark and is not used as a verifier.
Every model--benchmark pair uses all the prompts selected above.
A benchmark's overall score is the mean of its category scores, as GenEval and T2I-CompBench report it.

\paragraph{One sample.}
One sample is generated and scored under the protocol above, with T5 truncated to Flow-GRPO's 128 tokens and the official evaluator; every comparison between methods is made on the same base model, prompts and evaluator.

\clearpage
\subsection{Verifiers and Published Methods}
\label{sec:app_verifiers}

\paragraph{Verifiers.}
ImageReward and HPSv2.1 score every preview and final image.
VQAScore asks CLIP-FlanT5-XXL, the model its authors recommend, their question ``Does this figure show `\emph{prompt}'? Please answer yes or no.'' and takes the probability of ``Yes'' as the score (t2v\_metrics 3.0); it scores every preview and final image of the development candidates and, of the test candidates, every final image and the previews its two searches read (steps 8, 16 and 20).
PickScore and CLIP score, used as best-of-$N$ verifiers only, score the final images on SD3.5-Medium, and the step-8 previews and final images on Sana-0.6B.
CLIP score is 100 times the cosine similarity between the image and prompt embeddings of OpenAI's ViT-L/14, run with the QuickGELU activation its weights were trained with; PickScore is the released model's scaled similarity between image and prompt.

\paragraph{PSP.}
PSP~\cite{psp2026} halves its candidates at a quarter and at half of the trajectory; on the 40-step sampler half is step 20, and the quarter point, step 10, lies between the recorded steps 8 and 12.

\paragraph{FK steering.}
FK steering~\cite{singhal2025fk} is rerun as PSP~\cite{psp2026} reran it on SD3.5, with the resampling rule of its public code: four particles per image (160 NFE, the cost of best-of-4), the maximum ImageReward so far as potential with $\lambda=10$, multinomial resampling at the public code's step indices 8, 16, 24 and 32 of 40 (20, 40, 60 and 80\% of the trajectory; counted from zero, so after 9, 17, 25 and 33 model evaluations) and at the last step, and ImageReward read on each particle's clean-image estimate; the returned image is the particle whose final image ImageReward scores highest.
So that copies made at a resampling step diverge, every particle follows PSP's controlled-noise Euler sampler, which adds a small noise ($\gamma=0.005$, its SD3.5 value) after every deterministic Euler step of \cref{tab:main}'s sampler.
Each image starts its four particles from the initial noises of four of its eight candidates, so prompts and initial noises are those of \cref{tab:main}; FK steering with the \method score (\cref{sec:app_dropin}) keeps the same initial noises, random seeds and resampling steps.

\paragraph{An independent judge for the examples.}\label{sec:app_indep_judge}
The example figures from \cref{fig:app_mech_misread} on show a row only if GPT-5.4 (OpenAI, through AWS Bedrock), which none of the methods or verifiers compared in these figures uses, agrees with the outcome the row prints. It sees one final image with the prompt and answers whether the image matches the prompt exactly: every named object clearly present, in exactly the stated number, with the stated colour and in the stated spatial relation, ignoring style and image quality; for a row that shows one correct and one wrong image, it must answer yes for the first and no for the second. For T2I-CompBench and GenAI-Bench, whose rows print metric scores, it sees both images and says which matches the prompt more faithfully, once in each order, and must prefer \method's pick both times.

\clearpage
\subsection{Question Compilation}
\label{sec:app_questions}

Each slot of a GenEval prompt becomes one or more questions, asked about the preview or the final image in a single call together with the prompt.
\begin{itemize}
\item \emph{Presence}: ``Is there at least one \emph{X} in this image?'' (yes/no).
\item \emph{Count}: ``Are there exactly $n$ \emph{X}s in this image?'' and ``Are there $n{+}1$ or more \emph{X}s in this image?'', the second required to be false, following GenEval's rule for counting prompts; elsewhere ``at least $n$''.
\item \emph{Colour}: ``Is the \emph{X} \emph{c} in colour?'' (yes/no) and, for colour attribution, a choice among GenEval's ten colour names with ``there is no \emph{X}'' as a further option.
\item \emph{Position}: ``Is the \emph{X} \emph{left of} the \emph{Y}?'' (yes/no) and a box for each of the two objects.
\end{itemize}
The yes/no answers are combined by requiring every slot to hold, so the score of a candidate is the smallest of its slot probabilities.
For position, the boxes are fed to GenEval's own relation rule---the offset between box centres after subtracting a tenth of the object sizes, normalised---which we keep continuous through a logistic function centred at GenEval's threshold and multiply by the smaller presence probability; the position score is the mean of this geometric score and the image's smallest yes/no probability ($y$ in \cref{alg:app_score}).
For colour attribution, each object's score is the smaller of its presence probability and the probability of its requested colour.
These choices were made on development prompts (GenEval's generator with a different seed, no test prompt), rendered by Sana-0.6B, by comparing, per category, how well each way of asking orders candidates against the official verdicts.

\paragraph{Prompts without slots.}
GenEval ships the slots of every prompt; T2I-CompBench and GenAI-Bench give only the sentence.
For these, a language model (Claude Sonnet 4.6, temperature 0) writes the slots down once per prompt, before any image is generated, in GenEval's schema---objects with counts and colours, left/right/above/below relations---plus two slot types GenEval does not need: other adjectives of an object (shape, material, state; ``Is the \emph{X} \emph{a}?'') and other relations between two objects (``Is the \emph{X} \emph{r} the \emph{Y}?''), both yes/no.
The questions and the rule that combines them are those above; the benchmark's category picks the rule as GenEval's tag does (colour binding: the colour choice; spatial: the boxes; numeracy: exact counts; everything else: the smallest yes/no probability).
On GenEval's \compPrompts{} test prompts, whose slots are known, the same compiler reproduces every slot of \compExact{} prompts (\compExactPct\%) from the sentence alone---\compObjectPct\% of the objects, \compCountPct\% of the counts, \compColourPct\% of the colours and \compPositionPct\% of the relations; the other prompts spell COCO's ``hair drier'' as ``hair dryer'', add a colour the noun implies (an orange, a red stop sign) or, in one prompt, reverse a relation.

\clearpage
\subsection{Settings Fixed on Development Prompts}
\label{sec:app_settings}

For each judge, adaptive search takes the read step, number of candidates and threshold with the highest development accuracy at no more than the cost of best-of-2; reward-model thresholds are searched over quantiles of their scores, since their scales are not probabilities.
For the \method score this gives read step 16, six candidates and threshold 0.85: the first candidate is finished if its step-16 score reaches 0.85, otherwise the best of six.
VQAScore was tuned in the same way, with the read step restricted to those the test candidates record (8, 12, 16 and 20).
Pruning uses one schedule for every judge: eight candidates, the best four kept at step 8 and the best two at step 16, both finished and compared on their final images.
Qwen3-VL-8B's settings (\cref{sec:app_open_judge}) were chosen on the development prompts in the same way.

\paragraph{Other models and benchmarks.}
Every search setting is the one fixed on SD3.5-Medium's GenEval development prompts; nothing is tuned for another model or benchmark.
ImageReward and HPSv2.1 read every preview and final image and drive the same adaptive search and pruning as \method; PickScore, CLIP score and VQAScore score only the final images and serve as best-of-$N$ verifiers (on SD3.5-Medium's GenEval test prompts VQAScore also drives the two searches, \cref{tab:main}).

\clearpage
\subsection{Cost, Selection and Intervals}
\label{sec:app_cost}

Cost is the number of network evaluations per image, counting a guided evaluation once, as in the best-of-$N$ and search literature.
When several candidates share the highest score, the reported accuracy is the expectation over a uniformly random choice among them.
GenEval's overall score is the mean of its six category scores; each interval is the central 95\% of 2,000 paired bootstrap draws that resample prompts within each category.
\method's judge, the decision model Jev (\cref{sec:previews}), is called once per preview or final image.
Over the \nJevCalls{} calls that scored the test candidates, a call cost \jevCostPerCall{} on average, with a median latency of \jevLatency\,s; calls for the candidates of one image run in parallel.
Adaptive search reads a second batch of candidates for \adaptExtraPct\% of images, which gives \adaptCalls{} calls and about \adaptDollars{} per image; pruning makes \pruneCalls{} calls (eight previews at step 8, four at step 16, two finals), about \pruneDollars{} per image.

Intervals pooled over model--benchmark pairs average the pairs' differences and resample prompts within each category of each pair.

\clearpage
\subsection{Computation Resources}
\label{sec:app_compute}

Candidates were generated on NVIDIA RTX A6000 (48\,GB) and RTX PRO 6000 Blackwell (96\,GB) GPUs, and GenEval's official evaluator, the other benchmarks' metrics, the reward models, VQAScore, PickScore, CLIP score and Qwen3-VL-8B ran locally on the same machines.
Jev is called through its provider's API (cost and latency per call: \cref{sec:app_cost}); it adds no network evaluations to the generator, so every cost in NFE in the paper counts the generator alone.

\clearpage
\section{Further Results}
\label{sec:app_results}
\subsection{\texorpdfstring{\cref*{tab:main}}{Tab. 2} with Every Verifier}
\label{sec:app_main_full}

\noindent\begin{minipage}{\linewidth}\centering\captionsetup{type=table}
\centering
\footnotesize
\setlength{\tabcolsep}{4pt}
\begin{tabular}{lcccccc}
\toprule
Method & NFE & Single & Count. & Position & Col. attr. & Overall \\
\midrule
One sample & 40 & 0.99 & 0.56 & 0.26 & 0.58 & 0.68 \\
\midrule
Best-of-2, CLIP score & 80 & 0.99 & 0.63 & 0.28 & 0.61 & 0.71 \\
Best-of-2, PickScore & 80 & 1.00 & 0.67 & 0.28 & 0.62 & 0.72 \\
Best-of-2, VQAScore & 80 & 1.00 & 0.66 & 0.32 & 0.63 & 0.73 \\
Best-of-2, ImageReward & 80 & 1.00 & 0.64 & 0.30 & 0.64 & 0.72 \\
Best-of-2, HPSv2.1 & 80 & 1.00 & 0.66 & 0.28 & 0.62 & 0.72 \\
Best-of-2, \method score & 80 & 1.00 & 0.70 & 0.37 & 0.64 & 0.74 \\
Adaptive search, VQAScore & 71 & 1.00 & 0.71 & 0.39 & 0.64 & 0.75 \\
Adaptive search, ImageReward & 54 & 1.00 & 0.63 & 0.33 & 0.66 & 0.73 \\
Adaptive search, HPSv2.1 & 77 & 1.00 & 0.72 & 0.32 & 0.63 & 0.74 \\
\ours \method, adaptive (ours) & 73 & 1.00 & 0.79 & 0.53 & 0.69 & \textbf{0.80} \\
\midrule
Best-of-4, CLIP score & 160 & 1.00 & 0.68 & 0.29 & 0.64 & 0.73 \\
Best-of-4, PickScore & 160 & 1.00 & 0.74 & 0.28 & 0.66 & 0.75 \\
Best-of-4, VQAScore & 160 & 1.00 & 0.72 & 0.37 & 0.66 & 0.76 \\
Best-of-4, ImageReward & 160 & 1.00 & 0.69 & 0.32 & 0.69 & 0.75 \\
Best-of-4, HPSv2.1 & 160 & 1.00 & 0.72 & 0.29 & 0.65 & 0.74 \\
Best-of-4, \method score & 160 & 1.00 & 0.78 & 0.48 & 0.69 & 0.79 \\
PSP~\cite{psp2026}, ImageReward & 152 & 0.99 & 0.70 & 0.32 & 0.68 & 0.75 \\
FK steering~\cite{singhal2025fk}, ImageReward & 160 & 0.97 & 0.67 & 0.31 & 0.60 & 0.71 \\
Pruning, VQAScore & 144 & 0.99 & 0.78 & 0.35 & 0.68 & 0.77 \\
Pruning, ImageReward & 144 & 1.00 & 0.71 & 0.33 & 0.67 & 0.75 \\
Pruning, HPSv2.1 & 144 & 1.00 & 0.78 & 0.30 & 0.66 & 0.76 \\
\ours \method, pruning (ours) & 144 & 1.00 & 0.83 & 0.57 & 0.69 & \textbf{0.82} \\
\bottomrule
\end{tabular}

\caption{\textbf{\Cref{tab:main} with every verifier: each of \method's searches is significantly the most accurate overall in its block, by the intervals of \cref{tab:app_bon8}} (SD3.5-Medium, official test prompts; the same candidates and NFE). Overall is GenEval's score, the mean of all six categories, four of which are shown; bold: best overall accuracy in the block.}
\label{tab:app_main_full}\end{minipage}\par
\vspace{1.2em}

\Cref{tab:app_main_full} gives every verifier for \cref{tab:main}, which shows the strongest of them.

\clearpage
\subsection{Statistical Significance of the Main Results}
\label{sec:app_cat_ci}

\noindent\begin{minipage}{\linewidth}\centering\captionsetup{type=table}
\centering
\footnotesize
\setlength{\tabcolsep}{5pt}
\begin{tabular}{lcc}
\toprule
\method's search & \makecell{Adaptive search\\$-$ best-of-2} & \makecell{Pruning\\$-$ best-of-4} \\
\midrule
Single object & 0.000 \scriptsize[\textminus{}0.008, +0.006] & 0.000 \scriptsize[\textminus{}0.004, +0.002] \\
Two objects & \textbf{+0.029} \scriptsize[+0.004, +0.053] & +0.005 \scriptsize[\textminus{}0.011, +0.019] \\
Counting & \textbf{+0.098} \scriptsize[+0.061, +0.135] & \textbf{+0.044} \scriptsize[+0.022, +0.069] \\
Position & \textbf{+0.158} \scriptsize[+0.124, +0.194] & \textbf{+0.086} \scriptsize[+0.061, +0.111] \\
Colour attr. & \textbf{+0.043} \scriptsize[+0.011, +0.074] & +0.006 \scriptsize[\textminus{}0.020, +0.030] \\
\bottomrule
\end{tabular}

\caption{\textbf{The categories of \cref{fig:gaincat}: both searches gain significantly on counting and position, adaptive search also on two objects and colour attribution} (SD3.5-Medium, official test prompts): paired differences of \method's two searches against best-of-$N$, with 95\% bootstrap intervals over prompts; bold where the interval lies above zero. Best-of-$N$ uses the \method score on finished images at a higher cost (80 and 160 NFE against \adaptJevNfe{} and 144).}
\label{tab:app_cat_ci}\end{minipage}\par
\vspace{1.2em}

\Cref{tab:app_cat_ci} gives the categories of \cref{fig:gaincat} with their intervals: against best-of-$N$ with the \method score at a higher cost, adaptive search also gains on two-object and colour-attribution prompts.
\Cref{tab:app_bon8} gives the intervals of overall accuracy behind every comparison of \cref{tab:main} and the two gains printed in \cref{fig:compute}, all above zero, and those with best-of-8 and best-of-5: \method's adaptive search is above best-of-8 with each of the other five verifiers and matches best-of-5 with the \method score at well under half the cost, and its pruning matches best-of-8 with the \method score at under half the cost and is above it with every other verifier.

\begin{table}[!htbp]
\centering
\footnotesize
\setlength{\tabcolsep}{3pt}
\begin{tabular}{lcc}
\toprule
Minus & Adaptive search (\method) & Pruning (\method) \\
\midrule
\multicolumn{3}{l}{\emph{Best-of-2 / best-of-4 (the budgets of \cref{tab:main})}} \\
\method score & +0.054 \scriptsize[+0.043, +0.066] & +0.028 \scriptsize[+0.020, +0.036] \\
VQAScore & +0.071 \scriptsize[+0.058, +0.083] & +0.061 \scriptsize[+0.050, +0.072] \\
HPSv2.1 & +0.079 \scriptsize[+0.066, +0.091] & +0.076 \scriptsize[+0.062, +0.088] \\
ImageReward & +0.074 \scriptsize[+0.061, +0.087] & +0.068 \scriptsize[+0.055, +0.081] \\
PickScore & +0.079 \scriptsize[+0.066, +0.092] & +0.072 \scriptsize[+0.059, +0.085] \\
CLIP score & +0.090 \scriptsize[+0.077, +0.104] & +0.092 \scriptsize[+0.078, +0.104] \\
\midrule
\multicolumn{3}{l}{\emph{The same search driven by}} \\
VQAScore & +0.049 \scriptsize[+0.035, +0.062] & +0.053 \scriptsize[+0.039, +0.067] \\
HPSv2.1 & +0.063 \scriptsize[+0.048, +0.078] & +0.056 \scriptsize[+0.041, +0.071] \\
ImageReward & +0.073 \scriptsize[+0.058, +0.087] & +0.068 \scriptsize[+0.053, +0.083] \\
\midrule
PSP~\cite{psp2026}, ImageReward & -- & +0.068 \scriptsize[+0.053, +0.082] \\
FK steering~\cite{singhal2025fk}, ImageReward & +0.089 \scriptsize[+0.072, +0.107] & +0.110 \scriptsize[+0.091, +0.128] \\
\midrule
\multicolumn{3}{l}{\emph{Best-of-8 (320 NFE) and best-of-5 (200 NFE)}} \\
Best-of-8, \method score & -- & \textminus{}0.001 \scriptsize[\textminus{}0.010, +0.007] \\
Best-of-8, VQAScore & +0.018 \scriptsize[+0.003, +0.034] & +0.039 \scriptsize[+0.026, +0.052] \\
Best-of-8, HPSv2.1 & +0.044 \scriptsize[+0.027, +0.059] & +0.065 \scriptsize[+0.048, +0.081] \\
Best-of-8, ImageReward & +0.031 \scriptsize[+0.016, +0.047] & +0.052 \scriptsize[+0.038, +0.066] \\
Best-of-8, PickScore & +0.034 \scriptsize[+0.018, +0.050] & +0.055 \scriptsize[+0.039, +0.070] \\
Best-of-8, CLIP score & +0.058 \scriptsize[+0.040, +0.075] & +0.079 \scriptsize[+0.063, +0.094] \\
Best-of-5, \method score & \textminus{}0.004 \scriptsize[\textminus{}0.015, +0.006] & -- \\
\bottomrule
\end{tabular}

\caption{\textbf{\method's searches are significantly more accurate than every best-of-$N$ and search in their blocks of \cref{tab:main}, and than every other verifier's best-of-8 at under half its cost} (SD3.5-Medium, official test prompts; the searches of \cref{tab:main} at \adaptJevNfe{} and 144 NFE; PSP with its own schedule; FK steering with four particles, 160 NFE). Overall GenEval accuracy of \method's search minus that of the row's method; mean difference with its 95\% paired bootstrap interval over prompts resampled within category.}
\label{tab:app_bon8}
\end{table}

\clearpage
\subsection{Comparison with Published Methods}
\label{sec:app_published}

\begin{table}[!htbp]
\centering
\footnotesize
\setlength{\tabcolsep}{2.5pt}
\begin{tabular}{lccc}
\toprule
Method & Compute & GenEval & Gain \\
\midrule
\multicolumn{4}{l}{\emph{SD3.5-Large, as reported by PSP~\cite{psp2026}}} \\
One sample & 1.0$\times$ & 0.713 & -- \\
Best-of-$N$ & 4.0$\times$ & 0.747 & +0.034 \\
FK steering & 4.0$\times$ & 0.742 & +0.029 \\
DSearch & 4.0$\times$ & 0.704 & \textminus{}0.009 \\
NTS & 4.0$\times$ & 0.699 & \textminus{}0.014 \\
RBF & 4.0$\times$ & 0.738 & +0.025 \\
BFS & 4.0$\times$ & 0.747 & +0.034 \\
SVDD & 4.0$\times$ & 0.719 & +0.006 \\
PSP & 4.0$\times$ & 0.747 & +0.034 \\
\midrule
\multicolumn{4}{l}{\emph{SD3.5-Medium, this work}} \\
One sample & 1.0$\times$ & 0.678 & -- \\
Best-of-4, \method score & 4.0$\times$ & 0.791 & +0.113 \\
\ours \method, adaptive & 1.8$\times$ & 0.798 & \textbf{+0.120} \\
\ours \method, pruning & 3.6$\times$ & 0.819 & \textbf{+0.141} \\
\bottomrule
\end{tabular}

\caption{\textbf{Against published test-time methods: \method's searches gain more than best-of-$N$ at lower compute, which none of the seven published methods above does}. Upper block: as reported by PSP~\cite{psp2026} on SD3.5-Large (1024\,px, 32 steps). Lower block: \method on SD3.5-Medium (512\,px, 40 steps). Compute is a multiple of one sample; as models and samplers differ, only comparisons within each block (gains over one sample, and against best-of-$N$) are meaningful. Bold: \method's gains.}
\label{tab:published}
\end{table}

\Cref{tab:published} sets the gains PSP~\cite{psp2026} reports for seven test-time methods and best-of-$N$ on SD3.5-Large beside those of \method on SD3.5-Medium (\cref{sec:exp_search}); as models, resolutions and samplers differ, only comparisons within each block are meaningful: there, no published method exceeds best-of-$N$, whereas \method's pruning exceeds best-of-4, at lower compute.

\paragraph{PSP and FK steering rerun.}
With the quarter point at step 8 (\pspEightNfe{} NFE, as in \cref{tab:main}), PSP reaches \pspEightIR{} with ImageReward and \pspEightJev{} with the \method score, and \method's pruning schedule reaches \pruneJev{} at 144 NFE: it matches PSP's run with the \method score at lower cost, so the gain over PSP comes from the judge.
FK steering reaches \fkAcc{} overall; \method's pruning, at lower cost, exceeds it by \fkVsPrune{} and adaptive search, at under half its cost, by \fkVsAdapt{} (\cref{tab:app_bon8}).

\clearpage
\subsection{The \method Score as a Drop-In Verifier}
\label{sec:app_dropin}

\begin{figure}[!htbp]
\centering
\includegraphics[width=\linewidth]{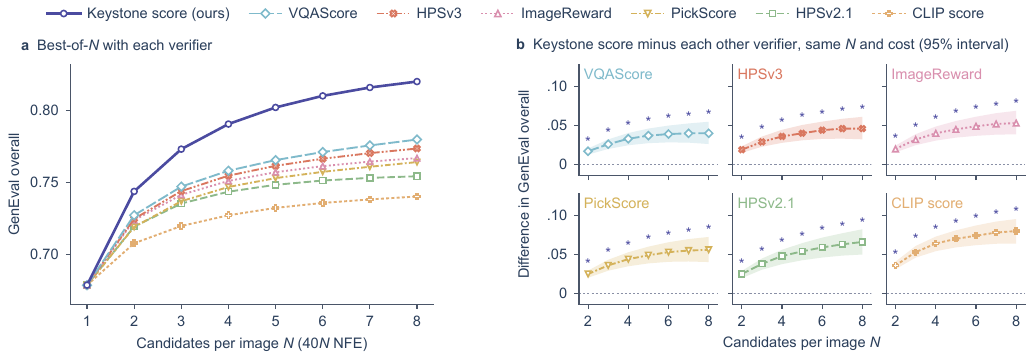}
\caption{\textbf{With the \method score as its verifier, best-of-$N$ is more accurate than with any other verifier at every $N$ from 2 to 8, and its lead widens from $N=2$ to $N=8$} (SD3.5-Medium, official test prompts, the candidates of \cref{tab:main}; correctness from GenEval's official evaluator). (a)~GenEval overall accuracy of best-of-$N$ on finished images with each verifier, from $N=1$, one sample, to $N=8$ (40$N$ NFE per image). (b)~Best-of-$N$ with the \method score minus best-of-$N$ with each other verifier at the same $N$ and cost: mean difference (line) and its 95\% paired bootstrap interval over prompts resampled within category (band); $*$: interval above zero, as for all 42 comparisons (smallest lower bound +0.013). Values: (a)~\cref{tab:app_bon_acc}; (b)~\cref{tab:app_dropin,tab:app_hpsv3}.}
\label{fig:app_trend_bon}
\end{figure}

\noindent\begin{minipage}{\linewidth}\centering\captionsetup{type=table}
\footnotesize
\setlength{\tabcolsep}{4pt}
\begin{tabular}{lccccccc}
\toprule
$N$ & \method score & VQAScore & ImageReward & HPSv2.1 & PickScore & CLIP score & HPSv3 \\
\midrule
1 & 0.678 & 0.678 & 0.678 & 0.678 & 0.678 & 0.678 & 0.678 \\
2 & \textbf{0.744} & 0.727 & 0.724 & 0.719 & 0.719 & 0.708 & 0.725 \\
3 & \textbf{0.773} & 0.747 & 0.741 & 0.735 & 0.737 & 0.720 & 0.744 \\
4 & \textbf{0.791} & 0.758 & 0.751 & 0.743 & 0.747 & 0.727 & 0.755 \\
5 & \textbf{0.802} & 0.765 & 0.757 & 0.748 & 0.753 & 0.732 & 0.762 \\
6 & \textbf{0.810} & 0.771 & 0.761 & 0.751 & 0.757 & 0.736 & 0.766 \\
7 & \textbf{0.816} & 0.776 & 0.764 & 0.753 & 0.761 & 0.738 & 0.770 \\
8 & \textbf{0.820} & 0.780 & 0.767 & 0.754 & 0.764 & 0.740 & 0.774 \\
\bottomrule
\end{tabular}

\vspace{0.6em}
\begin{tabular}{lccccc}
\toprule
$N$ & VQAScore & ImageReward & HPSv2.1 & PickScore & CLIP score \\
\midrule
2 & \makecell{+0.017\\\scriptsize[+0.013, +0.020]} & \makecell{+0.020\\\scriptsize[+0.016, +0.024]} & \makecell{+0.025\\\scriptsize[+0.020, +0.029]} & \makecell{+0.025\\\scriptsize[+0.020, +0.029]} & \makecell{+0.036\\\scriptsize[+0.031, +0.041]} \\
3 & \makecell{+0.026\\\scriptsize[+0.020, +0.032]} & \makecell{+0.032\\\scriptsize[+0.025, +0.039]} & \makecell{+0.038\\\scriptsize[+0.031, +0.045]} & \makecell{+0.036\\\scriptsize[+0.029, +0.043]} & \makecell{+0.053\\\scriptsize[+0.046, +0.061]} \\
4 & \makecell{+0.033\\\scriptsize[+0.025, +0.040]} & \makecell{+0.040\\\scriptsize[+0.031, +0.049]} & \makecell{+0.048\\\scriptsize[+0.038, +0.056]} & \makecell{+0.044\\\scriptsize[+0.035, +0.053]} & \makecell{+0.064\\\scriptsize[+0.054, +0.073]} \\
5 & \makecell{+0.037\\\scriptsize[+0.027, +0.046]} & \makecell{+0.045\\\scriptsize[+0.034, +0.056]} & \makecell{+0.054\\\scriptsize[+0.043, +0.064]} & \makecell{+0.049\\\scriptsize[+0.039, +0.060]} & \makecell{+0.070\\\scriptsize[+0.059, +0.081]} \\
6 & \makecell{+0.039\\\scriptsize[+0.028, +0.050]} & \makecell{+0.049\\\scriptsize[+0.037, +0.061]} & \makecell{+0.059\\\scriptsize[+0.046, +0.072]} & \makecell{+0.053\\\scriptsize[+0.041, +0.066]} & \makecell{+0.074\\\scriptsize[+0.062, +0.087]} \\
7 & \makecell{+0.040\\\scriptsize[+0.028, +0.053]} & \makecell{+0.052\\\scriptsize[+0.038, +0.066]} & \makecell{+0.063\\\scriptsize[+0.049, +0.077]} & \makecell{+0.055\\\scriptsize[+0.041, +0.070]} & \makecell{+0.078\\\scriptsize[+0.064, +0.092]} \\
8 & \makecell{+0.040\\\scriptsize[+0.026, +0.055]} & \makecell{+0.053\\\scriptsize[+0.039, +0.069]} & \makecell{+0.066\\\scriptsize[+0.050, +0.082]} & \makecell{+0.056\\\scriptsize[+0.040, +0.072]} & \makecell{+0.080\\\scriptsize[+0.064, +0.095]} \\
\bottomrule
\end{tabular}

\caption{\textbf{The \method score makes best-of-$N$ more accurate than any other verifier does at every $N$ from 2 to 8, with every paired interval above zero} (SD3.5-Medium, official test prompts, the candidates of \cref{tab:main}). Top: GenEval overall accuracy of best-of-$N$ on finished images with each verifier (40$N$ NFE per image), the values of \cref{fig:app_trend_bon}a; at $N=1$ every verifier keeps the one sample; bold: best in the row. Bottom: best-of-$N$ with the \method score minus best-of-$N$ with the column's verifier at the same $N$ and cost, with its 95\% paired bootstrap interval over prompts resampled within category, the values of \cref{fig:app_trend_bon}b.}
\label{tab:app_dropin}\label{tab:app_bon_acc}\end{minipage}\par
\vspace{1.2em}

\Cref{tab:app_dropin} keeps best-of-$N$ as it is and replaces only its verifier: at every $N$ from 2 to 8, best-of-$N$ with the \method score is more accurate than with each of the five other verifiers, and all 35 paired intervals lie above zero (smallest lower bound \dropinBonMinLo{}).
PSP~\cite{psp2026} improves the same way: with its schedule kept and the \method score in place of ImageReward, its own verifier, its accuracy rises by \dropinPspEightIr{} with the quarter point at step 8 and by \dropinPspTwelveIr{} with it at step 12.
So does FK steering~\cite{singhal2025fk}: with its four particles, resampling steps and random seeds kept and the \method score in place of ImageReward, its accuracy rises from \fkAcc{} to \fkjAcc{}, by \fkjOverFk{}.
\method's own searches still exceed FK steering with the \method score, at lower cost: pruning by \fkjVsPrune{} and adaptive search by \fkjVsAdapt{}; a better judge and searching at the key steps each contribute.

\clearpage
\subsection{Budget Allocation and the Oracle}
\label{sec:app_alloc}

In \cref{fig:adaptive} (left), the gap between the rates at which adaptive search draws further candidates for images whose first candidate is wrong and for those whose first is correct is \allocGapJev{} percentage points with \method, \allocGapHps{} with HPSv2.1 and \allocGapIr{} with ImageReward (95\% bootstrap intervals over prompts); \method's gap exceeds HPSv2.1's by \allocDiffHps{} and ImageReward's by \allocDiffIr{}.
Selecting among the eight candidates with GenEval's own detector, which is not available at test time, reaches \oracleOverall{} overall; it bounds what any judge can gain from these candidates.

\clearpage
\subsection{Intervals and Test Prompts for \texorpdfstring{\cref*{tab:judges}}{Tab. 1}}
\label{sec:app_judge_ci}

\begin{table}[!htbp]
\centering
\footnotesize
\setlength{\tabcolsep}{3pt}
\begin{minipage}[t]{0.485\linewidth}\centering
\begin{tabular}{llc}
\toprule
\multicolumn{3}{c}{\emph{Both read the step-8 preview}} \\
\midrule
Category & \method score minus & Difference \\
\midrule
Two objects & VQAScore & \makecell{\textbf{+0.16}\\\scriptsize[+0.08, +0.24]} \\
 & HPSv2.1 & \makecell{\textbf{+0.15}\\\scriptsize[+0.06, +0.24]} \\
 & ImageReward & \makecell{\textbf{+0.12}\\\scriptsize[+0.05, +0.19]} \\
\addlinespace[2pt]
Counting & VQAScore & \makecell{\textbf{+0.12}\\\scriptsize[+0.06, +0.18]} \\
 & HPSv2.1 & \makecell{\textbf{+0.20}\\\scriptsize[+0.14, +0.25]} \\
 & ImageReward & \makecell{\textbf{+0.34}\\\scriptsize[+0.24, +0.43]} \\
\addlinespace[2pt]
Colours & HPSv2.1 & \makecell{\textbf{+0.12}\\\scriptsize[+0.02, +0.19]} \\
 & ImageReward & \makecell{\textbf{+0.10}\\\scriptsize[+0.03, +0.18]} \\
\addlinespace[2pt]
Position & VQAScore & \makecell{\textbf{+0.30}\\\scriptsize[+0.23, +0.36]} \\
 & HPSv2.1 & \makecell{\textbf{+0.35}\\\scriptsize[+0.27, +0.42]} \\
 & ImageReward & \makecell{\textbf{+0.30}\\\scriptsize[+0.22, +0.37]} \\
\bottomrule
\end{tabular}
\end{minipage}\hfill
\begin{minipage}[t]{0.485\linewidth}\centering
\begin{tabular}{llc}
\toprule
\multicolumn{3}{c}{\emph{\method: step-8 preview; other judge: finished image}} \\
\midrule
Category & \method score minus & Difference \\
\midrule
Counting & HPSv2.1 & \makecell{\textbf{+0.09}\\\scriptsize[+0.01, +0.17]} \\
 & ImageReward & \makecell{\textbf{+0.14}\\\scriptsize[+0.06, +0.22]} \\
\addlinespace[2pt]
Position & VQAScore & \makecell{\textbf{+0.08}\\\scriptsize[+0.01, +0.14]} \\
 & HPSv2.1 & \makecell{\textbf{+0.35}\\\scriptsize[+0.26, +0.44]} \\
 & ImageReward & \makecell{\textbf{+0.21}\\\scriptsize[+0.13, +0.29]} \\
\bottomrule
\end{tabular}
\end{minipage}

\caption{\textbf{From the step-8 preview, the \method score orders pairs significantly better than the other judge in every comparison of \cref{tab:judges} stated in \cref{sec:exp_judges}} (SD3.5-Medium development prompts): pairs ordered correctly by the \method score minus those ordered by the other judge, with 95\% intervals from a paired bootstrap over prompts (2,000 draws); bold where the interval lies above zero. Left: both judges read the step-8 preview; right: the \method score reads the step-8 preview and the other judge the finished image.}
\label{tab:app_judge_ci}
\end{table}

\begin{table}[!htbp]
\centering
\footnotesize
\setlength{\tabcolsep}{3pt}
\begin{tabular}{lcccccc}
\toprule
 & \multicolumn{6}{c}{Preview at step 8} \\
\cmidrule(lr){2-7}
Judge & Single & Two obj. & Count. & Colours & Position & Col. attr. \\
\midrule
\ours \method score (ours) & \textbf{0.59} & \textbf{0.69} & \textbf{0.84} & \textbf{0.67} & \textbf{0.79} & \textbf{0.60} \\
HPSv2.1 & 0.49$^\dagger$ & 0.57$^\dagger$ & 0.73$^\dagger$ & 0.57$^\dagger$ & 0.53$^\dagger$ & 0.53$^\dagger$ \\
ImageReward & 0.45$^\dagger$ & 0.61$^\dagger$ & 0.61$^\dagger$ & 0.60$^\dagger$ & 0.54$^\dagger$ & 0.55$^\dagger$ \\
\bottomrule
\end{tabular}

\caption{\textbf{On the official test prompts, the \method score orders pairs from the same step-8 previews significantly better than both reward models, in every category} (SD3.5-Medium, 553 prompts $\times$ 4 images $\times$ 8 candidates; pairs within one image's candidates). Bold: best in the column. $^\dagger$: the \method score's step-8 advantage over that judge has a 95\% paired bootstrap interval above zero.}
\label{tab:app_judges_test}
\end{table}

\Cref{tab:app_judge_ci} gives the intervals behind the comparisons of \cref{tab:judges} stated in \cref{sec:exp_judges}. \Cref{tab:app_judges_test} repeats its step-8 reading against the two reward models, with every category, on the official test prompts: within each image's eight candidates, the step-8 preview's \method score orders the pairs with different verdicts better than the reward models do in every category, with paired intervals above zero (resampling prompts, each with its four images).

\clearpage
\subsection{Every Reading of \texorpdfstring{\cref*{fig:sana}}{Fig. 6}, with Qwen3-VL-8B's Intervals}
\label{sec:app_sana_judges}

\begin{table}[!htbp]
\centering
\footnotesize
\setlength{\tabcolsep}{2.5pt}
\begin{tabular*}{\linewidth}{@{\extracolsep{\fill}}llcccc}
\toprule
Judge & Pairs ordered & HPSv2.1 & ImageReward & PickScore & CLIP score \\
\midrule
Qwen3-VL-8B & Step-8 preview & \makecell{+0.06\\\scriptsize[+0.02, +0.10]$^\star$} & \makecell{+0.05\\\scriptsize[+0.01, +0.10]$^\star$} & \makecell{+0.08\\\scriptsize[+0.04, +0.11]$^\star$} & \makecell{+0.12\\\scriptsize[+0.07, +0.16]$^\star$} \\
Qwen3-VL-8B & Final image & \makecell{+0.06\\\scriptsize[+0.01, +0.10]$^\star$} & \makecell{+0.06\\\scriptsize[+0.01, +0.11]$^\star$} & \makecell{+0.08\\\scriptsize[+0.03, +0.12]$^\star$} & \makecell{+0.14\\\scriptsize[+0.10, +0.18]$^\star$} \\
Jev & Final image & \makecell{+0.06\\\scriptsize[+0.02, +0.11]$^\star$} & \makecell{+0.07\\\scriptsize[+0.02, +0.11]$^\star$} & \makecell{+0.08\\\scriptsize[+0.04, +0.12]$^\star$} & \makecell{+0.14\\\scriptsize[+0.10, +0.19]$^\star$} \\
\bottomrule
\end{tabular*}

\caption{\textbf{\method's questions answered by Qwen3-VL-8B order pairs significantly better than every preference model, from the step-8 preview and from the final image} (Sana-0.6B, GenEval counting and two-object, 179 prompts with four seeds each; the readings of \cref{fig:sana}). Pairs ordered correctly by \method's questions, answered by Qwen3-VL-8B or by Jev, minus those ordered by each other verifier reading the same image, with 95\% intervals from a paired bootstrap over prompts (2,000 draws); $^\star$: interval above zero.}
\label{tab:app_sana_ci}
\end{table}

\begin{table}[!htbp]
\centering
\footnotesize
\setlength{\tabcolsep}{2.4pt}
\begin{tabular}{lcccc}
\toprule
 & \multicolumn{2}{c}{Preview, step 8 of 20} & \multicolumn{2}{c}{Final image} \\
\cmidrule(lr){2-3}\cmidrule(lr){4-5}
Judge & Pairs ordered & Key-step pick & Pairs ordered & Best-of-6 \\
\midrule
\ours \method, Jev (ours) & \underline{0.682} & \underline{0.742} & \textbf{0.723} & \underline{0.745} \\
\ours \method, Qwen3-VL-8B (ours) & \textbf{0.701} & \textbf{0.744} & \underline{0.720} & \textbf{0.751} \\
HPSv2.1 & 0.641 & 0.735 & 0.661 & 0.728 \\
ImageReward & 0.647 & 0.723 & 0.658 & 0.725 \\
PickScore & 0.625 & 0.725 & 0.645 & 0.723 \\
CLIP score & 0.583 & 0.708 & 0.580 & 0.715 \\
\midrule
GenEval detector (reference) & 0.669 & 0.737 & -- & -- \\
\midrule
Single sample & -- & 0.681 & -- & 0.681 \\
Oracle & -- & 0.842 & -- & 0.842 \\
\bottomrule
\end{tabular}

\caption{\textbf{\Cref{fig:sana} as a table: \method's questions, answered by Jev or by Qwen3-VL-8B, are the two best of the six judges in every column} (Sana-0.6B; GenEval counting and two-object, 179 prompts $\times$ 4 seeds, six candidates that share the first two of 20 steps and branch there). Pairs ordered: fraction of candidate pairs with different official verdicts that the judge orders correctly, pooled over all pairs. Key-step pick and best-of-6: GenEval accuracy of the candidate the judge ranks first at step 8 and on final images. Bold: best; underlined: second best. The GenEval detector applied to the preview, single sample and oracle are references, not methods.}
\label{tab:app_sana}
\end{table}

\Cref{tab:app_sana} gives \cref{fig:sana} with every column. \Cref{tab:app_sana_ci} gives the paired bootstrap intervals over the 179 prompts: Qwen3-VL-8B asked \method's questions orders pairs better than HPSv2.1, ImageReward, PickScore and CLIP score, from the step-8 preview and from the final image, and Jev does so on the final image as well (all twelve intervals above zero).

\clearpage
\subsection{\method with an Open-Weight Judge}
\label{sec:app_open_judge}

The searches of \cref{tab:main} are repeated with an open judge, Qwen3-VL-8B-Instruct~\cite{bai2025qwen3vl}, under the protocol of \cref{sec:exp}.
It answers only the yes/no questions of \cref{sec:app_questions}, not Jev's colour choice or boxes; an image's score is its smallest slot probability, as for Jev.
It scores the candidates of \cref{tab:main}: \qwPrompts{} test prompts, \qwImages{} images of eight candidates each.
Pruning keeps that table's schedule.
Adaptive search's read step (8 or 16, the steps Qwen3-VL-8B scores), $K$ and $\tau$ were chosen on the development prompts by the rule and budget used for the reward models and VQAScore ($K$ from 2 to 8, $\tau$ at eight quantiles of the development scores): read step \qwR{}, $K=\qwK{}$, $\tau=\qwTau{}$.

On the test prompts (\cref{tab:qwen_search,tab:qwen_ci}), adaptive search with Qwen3-VL-8B lifts one sample's \qwOne{} to \qwAdapt{} at \qwAdaptNfe{} NFE and pruning to \qwPrune{} at 144 NFE.
Both are above best-of-$N$ at the same NFE with every verifier, including the \method score with Jev, and above the VQAScore, HPSv2.1 and ImageReward searches; every paired interval lies above zero.

\begin{table}[!t]
\centering
\footnotesize
\begin{tabular}{lrccc}
\toprule
Method & NFE & Count. & Position & Overall \\
\midrule
One sample & 40 & 0.56 & 0.26 & 0.678 \\
\midrule
Best-of-2, CLIP score & 80 & 0.63 & 0.28 & 0.708 \\
Best-of-2, PickScore & 80 & 0.67 & 0.28 & 0.719 \\
Best-of-2, VQAScore & 80 & 0.66 & 0.32 & 0.727 \\
Best-of-2, ImageReward & 80 & 0.64 & 0.30 & 0.724 \\
Best-of-2, HPSv2.1 & 80 & 0.66 & 0.28 & 0.719 \\
Best-of-2, \method, Jev & 80 & 0.70 & 0.37 & \textbf{0.744} \\
Best-of-2, \method, Qwen3-VL-8B & 80 & 0.69 & 0.36 & 0.741 \\
\midrule
Best-of-4, CLIP score & 160 & 0.68 & 0.29 & 0.727 \\
Best-of-4, PickScore & 160 & 0.74 & 0.28 & 0.747 \\
Best-of-4, VQAScore & 160 & 0.72 & 0.37 & 0.758 \\
Best-of-4, ImageReward & 160 & 0.69 & 0.32 & 0.751 \\
Best-of-4, HPSv2.1 & 160 & 0.72 & 0.29 & 0.743 \\
Best-of-4, \method, Jev & 160 & 0.78 & 0.48 & \textbf{0.791} \\
Best-of-4, \method, Qwen3-VL-8B & 160 & 0.79 & 0.45 & 0.785 \\
\midrule
Best-of-8, CLIP score & 320 & 0.72 & 0.28 & 0.740 \\
Best-of-8, PickScore & 320 & 0.79 & 0.29 & 0.764 \\
Best-of-8, VQAScore & 320 & 0.76 & 0.41 & 0.780 \\
Best-of-8, ImageReward & 320 & 0.71 & 0.34 & 0.767 \\
Best-of-8, HPSv2.1 & 320 & 0.73 & 0.29 & 0.754 \\
Best-of-8, \method, Jev & 320 & 0.83 & 0.58 & \textbf{0.820} \\
Best-of-8, \method, Qwen3-VL-8B & 320 & 0.86 & 0.55 & 0.819 \\
\midrule
Adaptive search, VQAScore & 71 & 0.71 & 0.39 & 0.749 \\
Adaptive search, ImageReward & 54 & 0.63 & 0.33 & 0.725 \\
Adaptive search, HPSv2.1 & 77 & 0.72 & 0.32 & 0.735 \\
\ours Adaptive search, \method, Jev & 73 & 0.79 & 0.53 & \textbf{0.798} \\
\ours Adaptive search, \method, Qwen3-VL-8B & 71 & 0.80 & 0.47 & 0.784 \\
\midrule
Pruning, VQAScore & 144 & 0.78 & 0.35 & 0.766 \\
Pruning, ImageReward & 144 & 0.71 & 0.33 & 0.751 \\
Pruning, HPSv2.1 & 144 & 0.78 & 0.30 & 0.763 \\
\ours Pruning, \method, Jev & 144 & 0.83 & 0.57 & \textbf{0.819} \\
\ours Pruning, \method, Qwen3-VL-8B & 144 & 0.85 & 0.48 & 0.808 \\
\bottomrule
\end{tabular}

\caption{\textbf{\method's searches with Qwen3-VL-8B as the judge: in every block, \method with either judge is more accurate overall than every other verifier} (SD3.5-Medium, GenEval). Rows are best-of-$N$ with each verifier and the two \method searches driven by each score, all on the candidates of \cref{tab:main}. NFE is network evaluations per image; overall is GenEval's score, the mean of its six category scores, given to three decimals, counting and position to two; bold: best overall in the block.}
\label{tab:qwen_search}
\end{table}

\begin{table}[!htbp]
\centering
\footnotesize
\begin{tabular}{lcc}
\toprule
\method with Qwen3-VL-8B minus & Adaptive search & Pruning \\
\midrule
Best-of-$N$ at the same NFE, \method, Qwen3-VL-8B & +0.058 \scriptsize[+0.046, +0.070] & +0.030 \scriptsize[+0.020, +0.039] \\
Best-of-$N$ at the same NFE, \method, Jev & +0.056 \scriptsize[+0.044, +0.067] & +0.024 \scriptsize[+0.013, +0.034] \\
Best-of-$N$ at the same NFE, HPSv2.1 & +0.074 \scriptsize[+0.062, +0.087] & +0.067 \scriptsize[+0.055, +0.079] \\
Best-of-$N$ at the same NFE, ImageReward & +0.071 \scriptsize[+0.059, +0.084] & +0.060 \scriptsize[+0.048, +0.073] \\
Best-of-$N$ at the same NFE, VQAScore & +0.068 \scriptsize[+0.057, +0.081] & +0.054 \scriptsize[+0.042, +0.066] \\
Best-of-$N$ at the same NFE, PickScore & +0.075 \scriptsize[+0.063, +0.087] & +0.065 \scriptsize[+0.053, +0.077] \\
Best-of-$N$ at the same NFE, CLIP score & +0.083 \scriptsize[+0.071, +0.096] & +0.083 \scriptsize[+0.070, +0.096] \\
\midrule
The same search, VQAScore & +0.035 \scriptsize[+0.021, +0.048] & +0.042 \scriptsize[+0.027, +0.056] \\
The same search, HPSv2.1 & +0.049 \scriptsize[+0.034, +0.064] & +0.045 \scriptsize[+0.032, +0.059] \\
The same search, ImageReward & +0.059 \scriptsize[+0.044, +0.073] & +0.057 \scriptsize[+0.041, +0.072] \\
\bottomrule
\end{tabular}

\caption{\textbf{With Qwen3-VL-8B as the judge, both \method searches are significantly more accurate than best-of-$N$ at the same NFE with every verifier, and than the same search with VQAScore, HPSv2.1 or ImageReward}. Overall GenEval accuracy on SD3.5-Medium; mean difference with its 95\% paired bootstrap interval over prompts resampled within category. Best-of-$N$ is taken at the same NFE, interpolated linearly between integer $N$.}
\label{tab:qwen_ci}
\end{table}

\clearpage
\subsection{Other Base Models and Benchmarks}
\label{sec:app_gen}

\begin{table}[!hbp]
\centering
\footnotesize
\setlength{\tabcolsep}{4pt}
\begin{tabular}{lcc>{\columncolor{oursbg}}cc>{\columncolor{oursbg}}c}
\toprule
 & One & \multicolumn{2}{c}{Adaptive budget} & \multicolumn{2}{c}{Pruning budget} \\
\cmidrule(lr){3-4}\cmidrule(lr){5-6}
Benchmark & sample & BoN-$k$ & \method & BoN-4 & \method \\
\midrule
GenEval (4 models) & 0.63 & 0.69 & \textbf{0.76}\rlap{$^\dagger$} & 0.71 & \textbf{0.77}\rlap{$^\dagger$} \\
T2I-CompBench (4 models) & 0.64 & 0.70 & \textbf{0.73}\rlap{$^\dagger$} & 0.71 & \textbf{0.74}\rlap{$^\dagger$} \\
\midrule
Mean of the 8 pairs & 0.64 & 0.69 & \textbf{0.74}\rlap{$^\dagger$} & 0.71 & \textbf{0.76}\rlap{$^\dagger$} \\
\midrule
GenAI-Bench (4 models) & 0.76 & 0.78 & \textbf{0.79}\rlap{$^\dagger$} & 0.78 & \textbf{0.80}\rlap{$^\dagger$} \\
\midrule
Mean of the 12 pairs & 0.68 & 0.72 & \textbf{0.76}\rlap{$^\dagger$} & 0.73 & \textbf{0.77}\rlap{$^\dagger$} \\
\bottomrule
\end{tabular}

\caption{\textbf{\Cref{fig:general} as a table, with GenAI-Bench: on every benchmark, averaged over the four base models, both of \method's searches are significantly above best-of-$N$ with the strongest other verifier} (benchmark score, the mean over SD3.5-Medium, FLUX.1-dev, SDXL and Sana-0.6B; GenAI-Bench is scored by VQAScore, so its strongest other verifier is among the other four; one search over eight candidates per prompt; 40 steps; settings fixed on SD3.5-Medium). BoN-$k$: best-of-$k$ with the strongest other verifier at the smallest $k$ that costs at least as much as \method's adaptive search; BoN-4 (160 NFE) likewise, against pruning (144 NFE). Adaptive search costs 71--103 NFE on GenEval and T2I-CompBench and 105--114 on GenAI-Bench ($k=2$ for SD3.5-Medium and FLUX.1-dev on GenEval, 3 otherwise), and the strongest other verifier is VQAScore on GenEval and T2I-CompBench, except ImageReward for Sana-0.6B on GenEval, and ImageReward on GenAI-Bench. Bold: the better of the two in each budget; $^\dagger$: above that best-of-$N$, pooled paired 95\% interval above zero.}
\label{tab:app_general}
\end{table}

\Cref{tab:app_general} gives \cref{fig:general} as a table.

\paragraph{Results.}
Averaged over the \genCells{} model--benchmark pairs, the differences from the strongest other verifier's best-of-$N$ are \genPoolAdapt{} \genPoolAdaptCI{} for adaptive search and \genPoolPrune{} \genPoolPruneCI{} for pruning on the 0--1 scale.
On GenAI-Bench, whose metric scores the image against the whole prompt, \method's searches are \genGbAbove{} best-of-$N$ with the strongest other verifier on average over its \genGbCells{} models with 300 prompts each (\genGbPoolAdapt{} \genGbPoolAdaptCI{} and \genGbPoolPrune{} \genGbPoolPruneCI{} on the 0--1 scale; \cref{tab:app_general}).
\paragraph{Examples.}\label{sec:app_gallery_other}
\Cref{fig:app_gallery_other_p1,fig:app_gallery_other_p2} show the base models and benchmarks of \cref{fig:general} at the pruning budget: \method's pruning (144 NFE) against best-of-4 (160 NFE) with the strongest other verifier, which draws the four leftmost candidates; FLUX.1-dev and SDXL on GenEval, Sana-0.6B on T2I-CompBench and SD3.5-Medium on GenAI-Bench are shown: together they cover all four base models and all three benchmarks, and on each \method's pruning is above that best-of-4 overall (SD3.5-Medium on GenEval is shown in \cref{fig:app_gallery_sd35_p1}). Each row shows a prompt's eight candidates as finished images with the \method scores of their step-8 previews (framed: the four kept) and step-16 previews (bold: the two finished), and both picks with the benchmark's verdict (GenEval's official evaluator) or score (T2I-CompBench: BLIP-VQA or UniDet by category; GenAI-Bench: VQAScore). Selection: a prompt qualifies when \method's pick is correct and best-of-4's wrong (GenEval) or scores at least 0.10 higher (the other two), with no tie at any cut; a row is shown only if GPT-5.4 (\cref{sec:app_indep_judge}) agrees: it calls \method's pick a match for the prompt and best-of-4's not (GenEval), or prefers \method's pick in both presentation orders (the other two). Each group shows its first three such prompts by prompt index.

\clearpage
\subsection{Readability on Every Base Model}
\label{sec:app_readgen}

\noindent\begin{minipage}{\linewidth}\centering\captionsetup{type=figure}
\centering
\includegraphics[width=\linewidth]{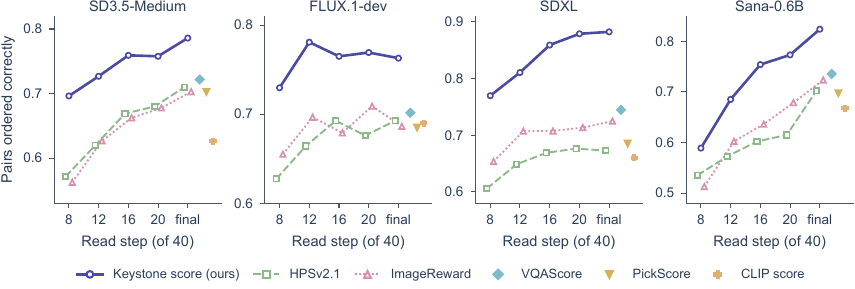}
\caption{\textbf{The key steps carry over: on every base model, \method reads previews better than the reward models read them} (GenEval, official test prompts, eight candidates per image). Pairs ordered correctly (y axis set per panel): the share of candidate pairs of one image with different official verdicts that the judge's score orders correctly (ties count half), averaged over GenEval's six categories. Lines: the \method score and the two reward models at every read step; points at ``final'': VQAScore, PickScore and CLIP score on finished images.}
\label{fig:readgen}\end{minipage}\par
\vspace{1.2em}

\Cref{fig:readgen} repeats the readability analysis of \cref{fig:keysteps} on the GenEval candidates of all four base models.
On each, the \method score orders previews better than HPSv2.1 and ImageReward order the same previews at every read step.
At step 16, \method's adaptive read step, the \method score reads previews significantly better than VQAScore (the best of the other judges on finished images) reads finished images, on \readgenSigModels{} (by \readgenSd{}, \readgenFlux{} and \readgenSdxl{}; bootstrap over prompts within category), so on these models the adaptive read step fixed on SD3.5-Medium falls where the outcome is already readable.
Against HPSv2.1 and ImageReward reading the same previews, the \method score is significantly better at every read step from 8 to 20 on all four base models (\readgenPrevSig{} of \readgenPrevAll{} paired intervals above zero, smallest lower bound \readgenPrevMinLo{}).

\clearpage
\subsection{Robustness to Search Settings}
\label{sec:app_robust}

\Cref{fig:robust} shows the robustness checks of \cref{sec:exp_analysis}: all eleven pruning schedules and every adaptive-search setting of the development grid, each with the \method score and with the two reward models, against best-of-$N$ at the same cost.
The \nSched{} schedules start $K_0$ candidates and keep the best $k$ at read step $r$, written $r{\to}k$: for $K_0=4$, $8{\to}1$, $12{\to}1$ and $16{\to}1$; for $K_0=6$, $12{\to}1$ and $8{\to}2$; for $K_0=8$, $8{\to}1$, $12{\to}1$, $8{\to}2$, $12{\to}2$, $8{\to}4$ then $12{\to}1$, and $8{\to}4$ then $16{\to}2$ (\cref{tab:main}'s).
The development grid of adaptive search has \nAdapt{} settings: read step 8, 12, 16 or 20; $K$ from 2 to 8; and nine thresholds---for the \method score 0.5, 0.6, 0.7, 0.8, 0.85, 0.9, 0.95 and 0.97, for the reward models eight quantiles of their development scores, and for every judge one threshold no score reaches, so that $K$ candidates are always drawn.
With the \method score, all \nSched{} pruning schedules are significantly above best-of-$N$ with the \method score, HPSv2.1 and ImageReward at the same cost (paired 95\% intervals above zero, best-of-$N$ interpolated linearly in NFE), and so are \adaptAboveAll{} adaptive settings. Of the \nSched{} schedules driven by HPSv2.1, \hpsAbove{} lie significantly above best-of-$N$ with HPSv2.1, and of those driven by ImageReward, \irAbove{} above best-of-$N$ with ImageReward.

\clearpage
\subsection{Scaling Past Eight Candidates}
\label{sec:app_scale}

Each test prompt has 32 candidates, which \cref{tab:main} uses as four independent searches over eight.
Taken as two searches over 16 and one over 32, they extend best-of-$N$ to $N=32$ for every verifier, and \method's pruning to 16 and 32 starting candidates under the rule of \cref{tab:main}'s schedule, which halves the candidates at steps 8 and 16 ($16{\to}8{\to}4$ and $32{\to}16{\to}8$; nothing is tuned; \cref{fig:scaling}).
Pruning from 32 candidates (\scalePruneThirtyTwoNfe{} NFE) is above best-of-32 (\scaleBonThirtyTwoNfe{} NFE) with each of the other five verifiers, with paired 95\% intervals above zero (smallest gain \scalePruneThirtyTwoMin{}, over \scalePruneThirtyTwoMinJudge{}).
From 16 candidates (\scalePruneSixteenNfe{} NFE), it is likewise above every other verifier's best-of-16 (\scaleBonSixteenNfe{} NFE; smallest gain \scalePruneSixteenMin{}, over \scalePruneSixteenMinJudge{}), and best-of-32 with the \method score is above best-of-32 with every other verifier (smallest gain \scaleBonMin{}, over \scaleBonMinJudge{}).

\clearpage
\subsection{Reading Final Images Instead of Previews}
\label{sec:app_readfinal}

The simplest search without previews finishes candidates one at a time and keeps the first whose final image reaches a threshold of the \method score (after $K$ candidates, the best of them).
With its threshold and $K$ picked on the development prompts within best-of-2's cost, as adaptive search's setting was ($\tau=\seqTwoTau$, $K=\seqTwoK$; \seqTwoNfe{} NFE on the test prompts), adaptive search, which reads step-16 previews, is above it (\seqTwoVsAdapt{}).
\Cref{fig:adaptive} (right) shows the same for adaptive search with the \method score, whose best setting within best-of-2's cost is least accurate when it reads final images.

\clearpage
\subsection{Image Quality of the Selected Images}
\label{sec:app_pick_quality}

\noindent\begin{minipage}{\linewidth}\centering\captionsetup{type=table}
\centering
\footnotesize
\setlength{\tabcolsep}{6pt}
\begin{tabular}{lcc}
\toprule
Picked by & HPSv2.1 & ImageReward \\
\midrule
One sample & 28.49 & 0.92 \\
\midrule
\ours \method, pruning & 29.23 \scriptsize(+0.74 [+0.65, +0.82]) & 1.13 \scriptsize(+0.21 [+0.18, +0.24]) \\
Best-of-8, \method score & 29.02 \scriptsize(+0.53 [+0.43, +0.61]) & 1.10 \scriptsize(+0.18 [+0.16, +0.21]) \\
\bottomrule
\end{tabular}

\caption{\textbf{The \method score's picks also score significantly higher than a random candidate on both reward models} (SD3.5-Medium, official test prompts). Mean HPSv2.1 and ImageReward of the final image each procedure picks; in brackets, the difference from one random candidate with its 95\% paired bootstrap interval over prompts. Pruning: the schedule of \cref{tab:main}, ties broken in a fixed order.}
\label{tab:app_pick_quality}\end{minipage}\par
\vspace{1.2em}

\Cref{tab:app_pick_quality} scores the final image each procedure picks with the two reward models: the picks of \method's pruning score higher than a random candidate on both, with paired intervals above zero.

\clearpage
\subsection{Compiled vs.\ Holistic Questions on Position}
\label{sec:app_ablcat}

\begin{table}[!htbp]
\centering
\footnotesize
\setlength{\tabcolsep}{5pt}
\begin{tabular}{lccc}
\toprule
 & Pairs ordered & \multicolumn{2}{c}{Pruning, GenEval} \\
\cmidrule(lr){3-4}
Judge & at step 8 & Overall & Position \\
\midrule
ImageReward & 0.56 & 0.75 & 0.33 \\
HPSv2.1 & 0.57 & 0.76 & 0.30 \\
One holistic question & 0.67 & 0.81 & 0.51 \\
\ours \method score (ours) & \textbf{0.70} & \textbf{0.82} & \textbf{0.57} \\
\bottomrule
\end{tabular}

\caption{\textbf{\Cref{fig:ablation} as a table: \method's compiled questions lead every column} (SD3.5-Medium, official test prompts). The same judge asked either \method's compiled questions or one holistic question, next to both reward models, all on the same candidates. Pairs ordered correctly at step 8: share of candidate pairs for the same image, one correct and one not, that the step-8 preview's score puts in the right order (ties count half), averaged over categories. Pruning: the schedule of \cref{tab:main} with each score.}
\label{tab:app_ablation}
\end{table}

On the candidates of \cref{fig:ablation} (\cref{tab:app_ablation}), with 95\% intervals from a paired bootstrap over prompts, Jev asked the one holistic question already orders step-8 preview pairs better than HPSv2.1 (\ablPairHolHps{}) and ImageReward (\ablPairHolIr{}), averaged over categories, and its pruning is more accurate than theirs (\ablPruneHolHps{} and \ablPruneHolIr{} overall); the compiled questions order step-8 pairs better than the holistic question (\ablPairCompHol{}).
On position, the compiled questions raise accuracy both when pruning (\ablPosPrune{}) and when picking among finished images with best-of-8 (\ablPosBonEight{}).
Position is the one slot where a compiled question also asks for boxes and applies GenEval's relation rule to them, which a single question about the whole prompt cannot do.

\clearpage
\subsection{Qualitative Examples}
\label{sec:app_qual}

\subsubsection{Selection of the \texorpdfstring{\cref*{fig:teaser}}{Fig. 1} Example}
\label{sec:app_teaser}

Among the development counting prompts whose eight candidates have mixed official verdicts, we keep those where the two candidates with the highest step-8 scores are both correct and the two with the lowest are both wrong, and show the one with the largest gap between the two pairs (the lower of the two highest scores minus the higher of the two lowest), with those four candidates.

\FloatBarrier
\subsubsection{What Pruning Reads on Two-Object and Colour Prompts}
\label{sec:app_qual_more}

\noindent\begin{minipage}{\linewidth}\centering\captionsetup{type=figure}
\centering
\includegraphics[width=\linewidth]{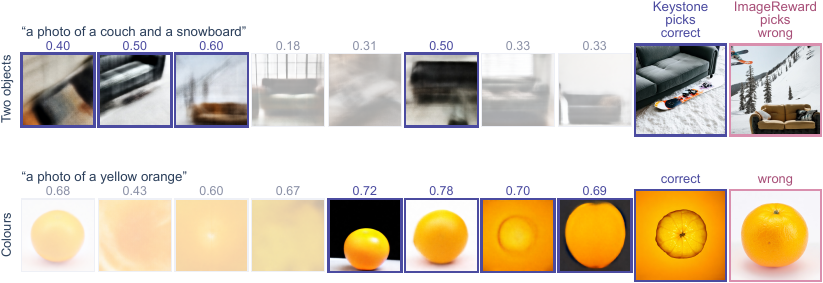}
\caption{\textbf{What pruning reads on two-object and colour prompts: \method's pruning ends correct where ImageReward's ends wrong} (SD3.5-Medium, official test prompts), by the rule of \cref{fig:qual}: among images where pruning with the \method score ends on a correct image and pruning with ImageReward on a wrong one, the one with the most wrong candidates.}
\label{fig:app_qual_more}\end{minipage}\par
\vspace{1.2em}

\Cref{fig:app_qual_more} applies the rule of \cref{fig:qual} to two-object and colour prompts.

\clearpage
\subsection{Comparison with HPSv3}
\label{sec:app_hpsv3}

\begin{table}[!htbp]
\centering
\footnotesize
\setlength{\tabcolsep}{4pt}
\begin{tabular}{lcccc}
\toprule
$N$ & NFE & HPSv3 & \method score & Difference \\
\midrule
2 & 80 & 0.725 & \textbf{0.744} & +0.019 \scriptsize[+0.015, +0.023] \\
3 & 120 & 0.744 & \textbf{0.773} & +0.029 \scriptsize[+0.023, +0.036] \\
4 & 160 & 0.755 & \textbf{0.791} & +0.036 \scriptsize[+0.028, +0.044] \\
5 & 200 & 0.762 & \textbf{0.802} & +0.040 \scriptsize[+0.031, +0.051] \\
6 & 240 & 0.766 & \textbf{0.810} & +0.044 \scriptsize[+0.033, +0.055] \\
7 & 280 & 0.770 & \textbf{0.816} & +0.046 \scriptsize[+0.033, +0.059] \\
8 & 320 & 0.774 & \textbf{0.820} & +0.046 \scriptsize[+0.032, +0.061] \\
\bottomrule
\end{tabular}

\caption{\textbf{The \method score is a significantly better verifier of best-of-$N$ than HPSv3 at every $N$ from 2 to 8} (SD3.5-Medium, official test prompts, the candidates of \cref{tab:main}). Overall GenEval accuracy of best-of-$N$ with each verifier at the same $N$ and cost, and the difference, \method score minus HPSv3, with its 95\% paired bootstrap interval over prompts resampled within category; bold: higher.}
\label{tab:app_hpsv3}
\end{table}

HPSv3~\cite{ma2025hpsv3} is a preference model built on a vision--language model and trained on human comparisons of a wide range of generated and real images; it is more recent than the five other verifiers the main text compares with.
We score the final image of every candidate of \cref{tab:main} with it and let it drive best-of-$N$ on the same candidates.
At every $N$ from 2 to 8, best-of-$N$ is more accurate with the \method score than with HPSv3, and all seven paired intervals lie above zero (smallest lower bound \hpsvBonMinLo{}; \cref{tab:app_hpsv3}).
\method's searches also exceed HPSv3's best-of-8 at under half its cost: adaptive search, at \hpsvAdaptNfe{} NFE, by \hpsvAdaptVsBonEight{}, and pruning, at 144 NFE, by \hpsvPruneVsBonEight{}.

\clearpage
\subsection{Comparisons at the Same Procedure and Cost}
\label{sec:app_sameproc}

\Cref{tab:app_sameproc} compares every row of \cref{tab:main} that uses the \method score with the same procedure, at the same cost, driven by each other verifier, and \method's pruning also with PSP and FK steering, which cost more.
All \sameprocAll{} differences, in the overall score and in the counting and position columns, lie significantly above zero (smallest lower bound \sameprocMinLo{}).
\Cref{tab:app_sameproc_scale} does the same for \cref{fig:scaling}: pruning from $K_0$ candidates is significantly more accurate than best-of-$K_0$ with every other verifier for $K_0 = 8$, 16 and 32 (smallest lower bound \scaleSameKMinLo{}), at well under its cost.
For \cref{fig:robust}, each of the \robustSchedules{} pruning schedules is significantly more accurate with the \method score than the same schedule with HPSv2.1 or with ImageReward (all \robustSameAll{} intervals above zero, smallest lower bound \robustSameMinLo{}).

\begin{table}[!htbp]
\centering
\footnotesize
\setlength{\tabcolsep}{4pt}
\begin{tabular}{llccc}
\toprule
Row of \cref{tab:main} & minus the same procedure with & Overall & Count. & Position \\
\midrule
Best-of-2, \method score & VQAScore & \makecell{\textbf{+0.017}\\\scriptsize[+0.013, +0.020]} & \makecell{\textbf{+0.031}\\\scriptsize[+0.019, +0.043]} & \makecell{\textbf{+0.049}\\\scriptsize[+0.037, +0.061]} \\
 & HPSv2.1 & \makecell{\textbf{+0.025}\\\scriptsize[+0.020, +0.029]} & \makecell{\textbf{+0.035}\\\scriptsize[+0.023, +0.046]} & \makecell{\textbf{+0.089}\\\scriptsize[+0.073, +0.106]} \\
 & ImageReward & \makecell{\textbf{+0.020}\\\scriptsize[+0.016, +0.024]} & \makecell{\textbf{+0.054}\\\scriptsize[+0.041, +0.068]} & \makecell{\textbf{+0.071}\\\scriptsize[+0.057, +0.087]} \\
 & PickScore & \makecell{\textbf{+0.025}\\\scriptsize[+0.020, +0.029]} & \makecell{\textbf{+0.029}\\\scriptsize[+0.019, +0.040]} & \makecell{\textbf{+0.092}\\\scriptsize[+0.076, +0.109]} \\
 & CLIP score & \makecell{\textbf{+0.036}\\\scriptsize[+0.031, +0.041]} & \makecell{\textbf{+0.067}\\\scriptsize[+0.054, +0.080]} & \makecell{\textbf{+0.087}\\\scriptsize[+0.070, +0.102]} \\
\midrule
\method, adaptive & adaptive search, VQAScore & \makecell{\textbf{+0.049}\\\scriptsize[+0.035, +0.062]} & \makecell{\textbf{+0.081}\\\scriptsize[+0.033, +0.133]} & \makecell{\textbf{+0.140}\\\scriptsize[+0.100, +0.180]} \\
 & adaptive search, HPSv2.1 & \makecell{\textbf{+0.063}\\\scriptsize[+0.048, +0.078]} & \makecell{\textbf{+0.078}\\\scriptsize[+0.027, +0.135]} & \makecell{\textbf{+0.212}\\\scriptsize[+0.165, +0.263]} \\
 & adaptive search, ImageReward & \makecell{\textbf{+0.073}\\\scriptsize[+0.058, +0.087]} & \makecell{\textbf{+0.163}\\\scriptsize[+0.113, +0.218]} & \makecell{\textbf{+0.200}\\\scriptsize[+0.152, +0.250]} \\
\midrule
Best-of-4, \method score & VQAScore & \makecell{\textbf{+0.033}\\\scriptsize[+0.025, +0.040]} & \makecell{\textbf{+0.062}\\\scriptsize[+0.039, +0.084]} & \makecell{\textbf{+0.110}\\\scriptsize[+0.082, +0.139]} \\
 & HPSv2.1 & \makecell{\textbf{+0.048}\\\scriptsize[+0.038, +0.056]} & \makecell{\textbf{+0.065}\\\scriptsize[+0.038, +0.093]} & \makecell{\textbf{+0.190}\\\scriptsize[+0.155, +0.224]} \\
 & ImageReward & \makecell{\textbf{+0.040}\\\scriptsize[+0.031, +0.049]} & \makecell{\textbf{+0.093}\\\scriptsize[+0.062, +0.122]} & \makecell{\textbf{+0.158}\\\scriptsize[+0.129, +0.191]} \\
 & PickScore & \makecell{\textbf{+0.044}\\\scriptsize[+0.035, +0.053]} & \makecell{\textbf{+0.044}\\\scriptsize[+0.022, +0.066]} & \makecell{\textbf{+0.197}\\\scriptsize[+0.163, +0.233]} \\
 & CLIP score & \makecell{\textbf{+0.064}\\\scriptsize[+0.054, +0.073]} & \makecell{\textbf{+0.105}\\\scriptsize[+0.078, +0.134]} & \makecell{\textbf{+0.190}\\\scriptsize[+0.157, +0.224]} \\
\midrule
\method, pruning & pruning, VQAScore & \makecell{\textbf{+0.053}\\\scriptsize[+0.039, +0.067]} & \makecell{\textbf{+0.050}\\\scriptsize[+0.012, +0.091]} & \makecell{\textbf{+0.220}\\\scriptsize[+0.168, +0.273]} \\
 & pruning, HPSv2.1 & \makecell{\textbf{+0.056}\\\scriptsize[+0.041, +0.071]} & \makecell{\textbf{+0.044}\\\scriptsize[+0.007, +0.088]} & \makecell{\textbf{+0.263}\\\scriptsize[+0.207, +0.323]} \\
 & pruning, ImageReward & \makecell{\textbf{+0.068}\\\scriptsize[+0.053, +0.083]} & \makecell{\textbf{+0.116}\\\scriptsize[+0.066, +0.166]} & \makecell{\textbf{+0.242}\\\scriptsize[+0.190, +0.297]} \\
 & PSP, ImageReward & \makecell{\textbf{+0.068}\\\scriptsize[+0.053, +0.082]} & \makecell{\textbf{+0.125}\\\scriptsize[+0.074, +0.182]} & \makecell{\textbf{+0.247}\\\scriptsize[+0.195, +0.302]} \\
 & FK steering, ImageReward & \makecell{\textbf{+0.110}\\\scriptsize[+0.091, +0.128]} & \makecell{\textbf{+0.163}\\\scriptsize[+0.106, +0.220]} & \makecell{\textbf{+0.260}\\\scriptsize[+0.205, +0.315]} \\
\bottomrule
\end{tabular}

\caption{\textbf{Every row of \cref{tab:main} with the \method score is significantly above its procedure with every other verifier, and pruning also above PSP and FK steering, overall and on counting and position} (SD3.5-Medium, official test prompts). Mean difference with its 95\% paired bootstrap interval over prompts; bold: above zero.}
\label{tab:app_sameproc}
\end{table}

\begin{table}[!htbp]
\centering
\footnotesize
\setlength{\tabcolsep}{4pt}
\begin{tabular}{lccccc}
\toprule
\method's pruning from & \multicolumn{5}{c}{minus best-of-$K_0$ with each verifier} \\
\cmidrule(lr){2-6}
 & VQAScore & HPSv2.1 & ImageReward & PickScore & CLIP score \\
\midrule
$K_0=8$ & \makecell{\textbf{+0.039}\\\scriptsize[+0.026, +0.052]} & \makecell{\textbf{+0.065}\\\scriptsize[+0.048, +0.081]} & \makecell{\textbf{+0.052}\\\scriptsize[+0.038, +0.066]} & \makecell{\textbf{+0.055}\\\scriptsize[+0.039, +0.070]} & \makecell{\textbf{+0.079}\\\scriptsize[+0.063, +0.094]} \\
$K_0=16$ & \makecell{\textbf{+0.048}\\\scriptsize[+0.029, +0.069]} & \makecell{\textbf{+0.075}\\\scriptsize[+0.054, +0.096]} & \makecell{\textbf{+0.068}\\\scriptsize[+0.047, +0.090]} & \makecell{\textbf{+0.081}\\\scriptsize[+0.058, +0.103]} & \makecell{\textbf{+0.095}\\\scriptsize[+0.074, +0.117]} \\
$K_0=32$ & \makecell{\textbf{+0.048}\\\scriptsize[+0.019, +0.078]} & \makecell{\textbf{+0.063}\\\scriptsize[+0.032, +0.092]} & \makecell{\textbf{+0.069}\\\scriptsize[+0.038, +0.101]} & \makecell{\textbf{+0.089}\\\scriptsize[+0.057, +0.120]} & \makecell{\textbf{+0.090}\\\scriptsize[+0.061, +0.122]} \\
\bottomrule
\end{tabular}

\caption{\textbf{\method's pruning from $K_0$ candidates beats best-of-$K_0$ with every other verifier} (SD3.5-Medium, official test prompts, the candidates of \cref{fig:scaling}). Overall GenEval accuracy, pruning minus best-of-$K_0$, with its 95\% paired bootstrap interval over prompts; bold: interval above zero.}
\label{tab:app_sameproc_scale}
\end{table}

\clearpage
\subsection{A Consistent Lead across Categories, Read Steps and Judges}
\label{sec:app_nonsig}

\begin{figure}[!hbp]
\centering
\includegraphics[width=\linewidth]{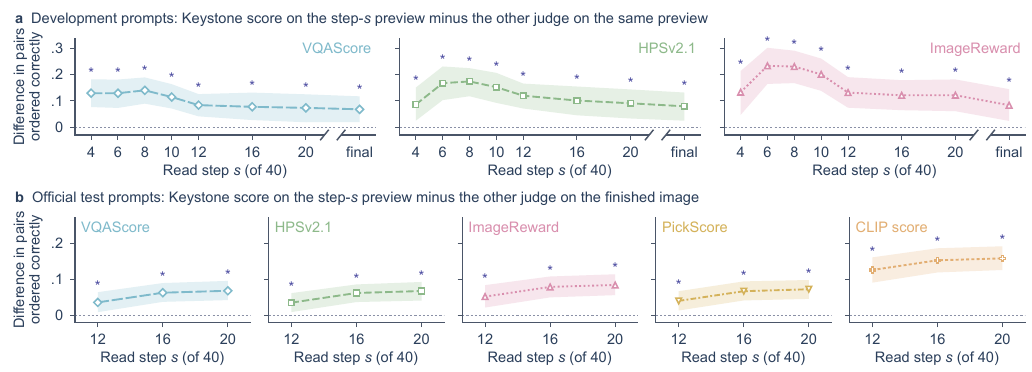}
\caption{\textbf{From step 4 on, the \method score orders candidates better than VQAScore and both reward models reading the same image, and from step 12 its previews order them better than every other judge orders the finished images} (SD3.5-Medium, counting and two-object prompts). (a)~Development prompts, the same preview; (b)~official test prompts, against the finished image. Band: 95\% paired bootstrap interval over prompts; $*$: interval above zero, as for all 39 comparisons. Values: the bottom block of \cref{tab:app_nonsig}.}
\label{fig:app_trend_readstep}
\end{figure}

\begin{table}[!t]
\centering
\footnotesize
\setlength{\tabcolsep}{2.5pt}
\begin{tabular*}{\linewidth}{@{\extracolsep{\fill}}llcc}
\toprule
\cref{tab:judges}, step-8 reading & \method score minus & Development prompts & Test prompts \\
\midrule
Counting, preview against the final image & VQAScore & \makecell{+0.034\\\scriptsize[\textminus{}0.019, +0.092]} & \makecell{+0.055\\\scriptsize[+0.023, +0.087]$^\star$} \\
Colours & VQAScore & \makecell{+0.059\\\scriptsize[\textminus{}0.048, +0.168]} & \makecell{+0.092\\\scriptsize[+0.033, +0.143]$^\star$} \\
Colour attribution & HPSv2.1 & \makecell{+0.055\\\scriptsize[\textminus{}0.021, +0.133]} & \makecell{+0.069\\\scriptsize[+0.034, +0.106]$^\star$} \\
\bottomrule
\end{tabular*}

\vspace{0.8em}
\begin{tabular*}{\linewidth}{@{\extracolsep{\fill}}lccccc}
\toprule
Read step & VQAScore & HPSv2.1 & ImageReward & PickScore & CLIP score \\
\midrule
\multicolumn{6}{l}{\makebox[0pt][l]{\emph{Development prompts (counting and two objects): \method's preview minus the other judge on the same preview}}} \\
Step 4 & \makecell{+0.129\\\scriptsize[+0.076, +0.181]$^\star$} & \makecell{+0.086\\\scriptsize[+0.024, +0.150]$^\star$} & \makecell{+0.133\\\scriptsize[+0.046, +0.213]$^\star$} & -- & -- \\
Step 6 & \makecell{+0.129\\\scriptsize[+0.073, +0.181]$^\star$} & \makecell{+0.167\\\scriptsize[+0.103, +0.231]$^\star$} & \makecell{+0.233\\\scriptsize[+0.164, +0.301]$^\star$} & -- & -- \\
Step 8 & \makecell{+0.140\\\scriptsize[+0.089, +0.189]$^\star$} & \makecell{+0.174\\\scriptsize[+0.117, +0.224]$^\star$} & \makecell{+0.230\\\scriptsize[+0.167, +0.291]$^\star$} & -- & -- \\
Step 10 & \makecell{+0.114\\\scriptsize[+0.071, +0.163]$^\star$} & \makecell{+0.152\\\scriptsize[+0.092, +0.206]$^\star$} & \makecell{+0.200\\\scriptsize[+0.138, +0.261]$^\star$} & -- & -- \\
Step 12 & \makecell{+0.084\\\scriptsize[+0.041, +0.126]$^\star$} & \makecell{+0.119\\\scriptsize[+0.071, +0.165]$^\star$} & \makecell{+0.131\\\scriptsize[+0.074, +0.190]$^\star$} & -- & -- \\
Step 16 & \makecell{+0.077\\\scriptsize[+0.027, +0.131]$^\star$} & \makecell{+0.101\\\scriptsize[+0.045, +0.154]$^\star$} & \makecell{+0.121\\\scriptsize[+0.064, +0.178]$^\star$} & -- & -- \\
Step 20 & \makecell{+0.073\\\scriptsize[+0.018, +0.125]$^\star$} & \makecell{+0.091\\\scriptsize[+0.033, +0.143]$^\star$} & \makecell{+0.121\\\scriptsize[+0.061, +0.181]$^\star$} & -- & -- \\
Final image & \makecell{+0.068\\\scriptsize[+0.019, +0.117]$^\star$} & \makecell{+0.079\\\scriptsize[+0.024, +0.131]$^\star$} & \makecell{+0.083\\\scriptsize[+0.024, +0.144]$^\star$} & -- & -- \\
\midrule
\multicolumn{6}{l}{\makebox[0pt][l]{\emph{Official test prompts (counting and two objects): \method's preview minus the other judge on the finished image}}} \\
Step 12 & \makecell{+0.036\\\scriptsize[+0.009, +0.064]$^\star$} & \makecell{+0.035\\\scriptsize[+0.009, +0.062]$^\star$} & \makecell{+0.052\\\scriptsize[+0.022, +0.084]$^\star$} & \makecell{+0.040\\\scriptsize[+0.013, +0.067]$^\star$} & \makecell{+0.126\\\scriptsize[+0.091, +0.161]$^\star$} \\
Step 16 & \makecell{+0.063\\\scriptsize[+0.037, +0.090]$^\star$} & \makecell{+0.062\\\scriptsize[+0.037, +0.086]$^\star$} & \makecell{+0.079\\\scriptsize[+0.050, +0.108]$^\star$} & \makecell{+0.067\\\scriptsize[+0.041, +0.094]$^\star$} & \makecell{+0.153\\\scriptsize[+0.119, +0.186]$^\star$} \\
Step 20 & \makecell{+0.068\\\scriptsize[+0.042, +0.096]$^\star$} & \makecell{+0.068\\\scriptsize[+0.041, +0.093]$^\star$} & \makecell{+0.084\\\scriptsize[+0.056, +0.114]$^\star$} & \makecell{+0.072\\\scriptsize[+0.046, +0.098]$^\star$} & \makecell{+0.158\\\scriptsize[+0.126, +0.192]$^\star$} \\
\bottomrule
\end{tabular*}

\caption{\textbf{The \method score leads in every comparison across categories and read steps, with every interval above zero on the official test prompts and in every comparison of \cref{fig:keysteps}} (SD3.5-Medium; paired bootstrap over prompts). Top: step-8 readings of \cref{tab:judges}. Bottom: the values of \cref{fig:app_trend_readstep}. Difference: the \method score minus the other judge, with its 95\% interval; $^\star$: above zero.}
\label{tab:app_nonsig}
\end{table}

Beyond the comparisons of the main text, the \method score leads wherever the same readings are taken on more prompts, at more read steps or with other judges; \cref{tab:app_nonsig} gives each comparison with its interval.

\paragraph{Categories on the test prompts.}
On the official test prompts, which hold four images per prompt, the \method score reads step-8 previews significantly better than both reward models in every category (\cref{tab:app_judges_test}); it also reads colours significantly better than VQAScore (\supTestColVqa{}), and counts from the step-8 preview significantly better than VQAScore reads them from finished images (\supTestCountVqaCross{}).

\paragraph{Previews against previews and against finished images.}
On the development prompts of \cref{fig:keysteps}, the \method score orders candidate pairs from its previews significantly better than VQAScore, HPSv2.1 and ImageReward order the same previews, at every read step from 4 to the final image; on the official test prompts, its previews from step 12 on, after 30\% of the generation, order them significantly better than every other judge orders the finished images (\cref{fig:app_trend_readstep}, \cref{tab:app_nonsig}).

\clearpage
\subsection{How the Outcome Is Decided and Read, and More Examples}
\label{sec:app_mech}\label{sec:app_gallery}

\begin{figure}[!h]
\centering
\includegraphics[width=\linewidth]{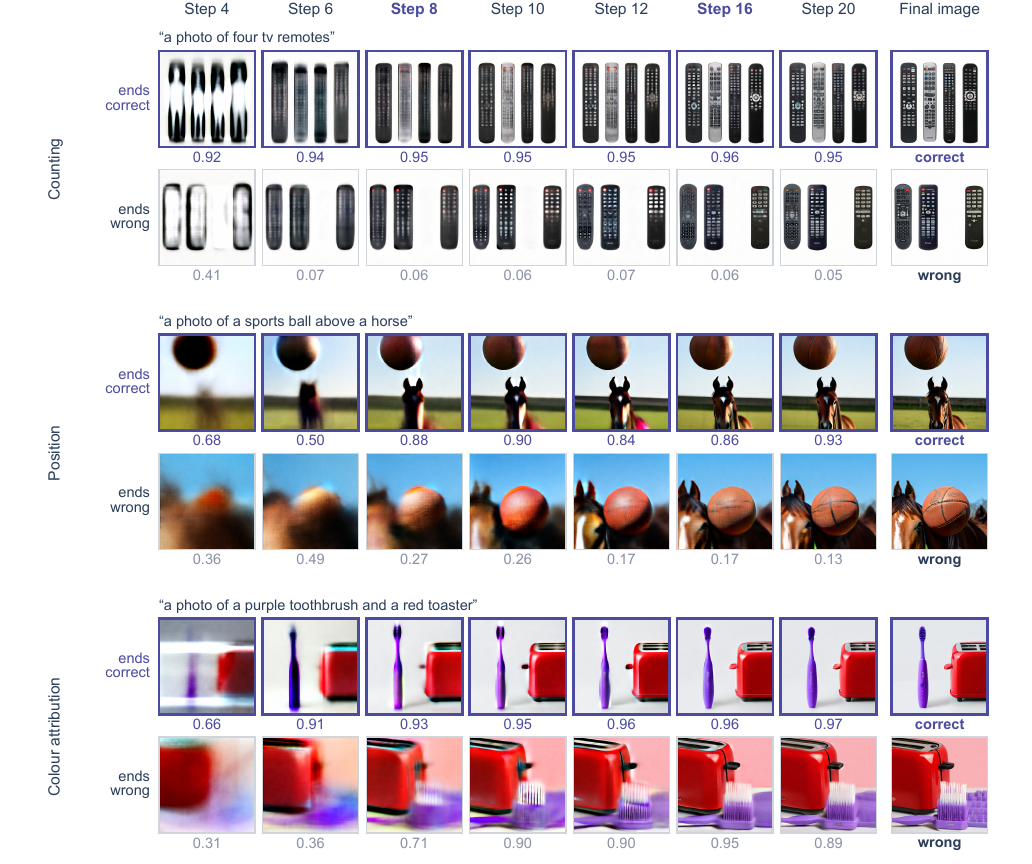}
\caption{\textbf{In each example, \method's score ranks the candidate that ends correct above the one that ends wrong at every preview from step 4 on} (SD3.5-Medium, development prompts). Each pair of rows: two candidates of one prompt, their previews with the \method score under each and the final image with the verdict of GenEval's official evaluator; indigo heads: pruning's read steps. Selection: per category, the first development prompt whose eight candidates have mixed verdicts, with its highest-scoring correct and lowest-scoring wrong candidate at step 8 (of 41, 56 and 73 such prompts).}
\label{fig:app_mech_sequence}
\end{figure}

\begin{figure}[p]
\centering
\includegraphics[width=\linewidth]{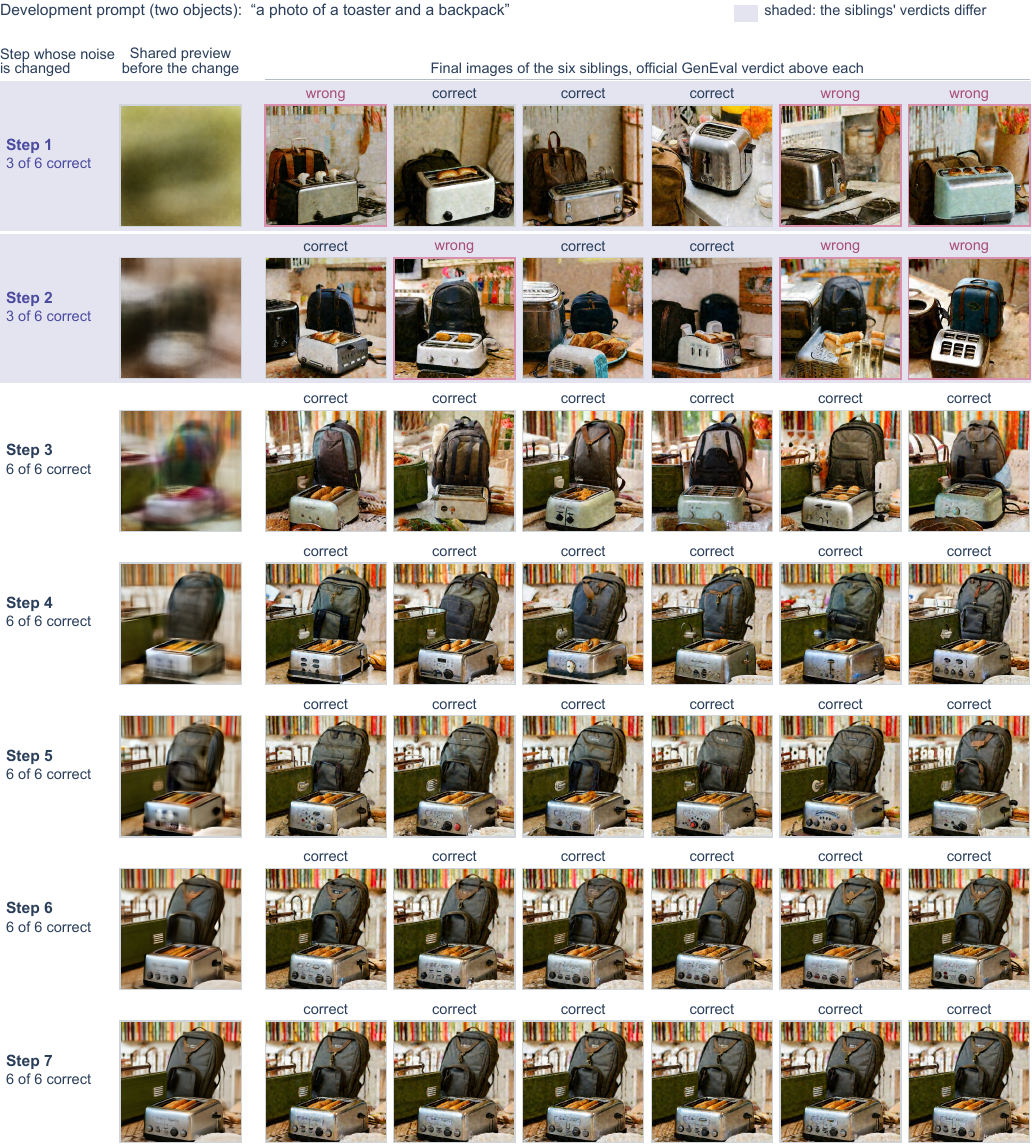}
\caption{\textbf{Here, a new noise draw at the first step leaves three of the six siblings correct and three wrong, and so does one at step 2; a new draw at any of steps 3--7 leaves all six correct} (SD3.5-Medium, development prompt, the 10-step stochastic sampler of \cref{fig:keysteps}). Each row is one sibling group of \cref{fig:keysteps} (left): six siblings share the starting noise and every noise draw except the one at the row's step. Left: the preview they share just before that step; right: their final images with the verdict of GenEval's official evaluator; shaded: the verdicts differ. Over all 40 prompt--seed pairs of \cref{fig:keysteps} (left), the verdicts differ in 29 and 26 of the 40 groups at steps 1 and 2 and in at most 17 at each of steps 3--7. Selection: 9 of the 40 pairs split three to three at step 1; shown is the first of them in prompt order.}
\label{fig:app_mech_branch}
\end{figure}

\begin{figure}[p]
\centering
\includegraphics[width=\linewidth]{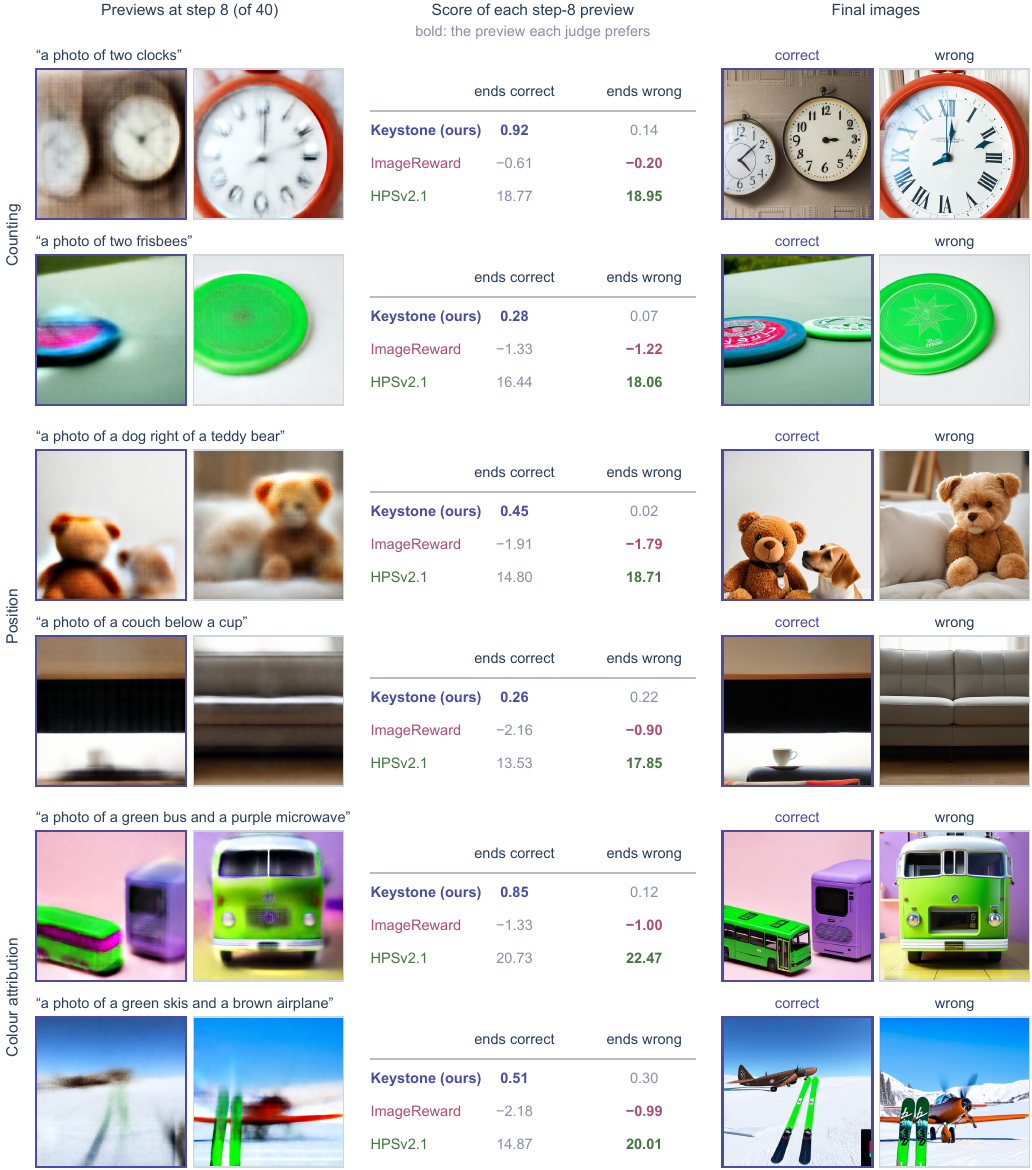}
\caption{\textbf{The same step-8 previews read by three judges: \method's questions prefer the candidate that ends correct, ImageReward and HPSv2.1 the one that ends wrong} (SD3.5-Medium, test prompts). Each row: two candidates of one image, with step-8 previews (framed: \method's preference), each judge's scores (bold: its preference) and final images with GenEval's official verdict. Selection: of the 364, 647 and 578 candidate pairs (counting, position, colour attribution) with a correct and a wrong candidate of one image that the \method score at step 8 orders correctly and both reward models the other way, strictly, the first two prompts per category whose verdicts GPT-5.4 confirms (\cref{sec:app_indep_judge}); per prompt, its first such image and pair.}
\label{fig:app_mech_misread}
\end{figure}

\begin{figure}[p]
\centering
\includegraphics[width=\linewidth]{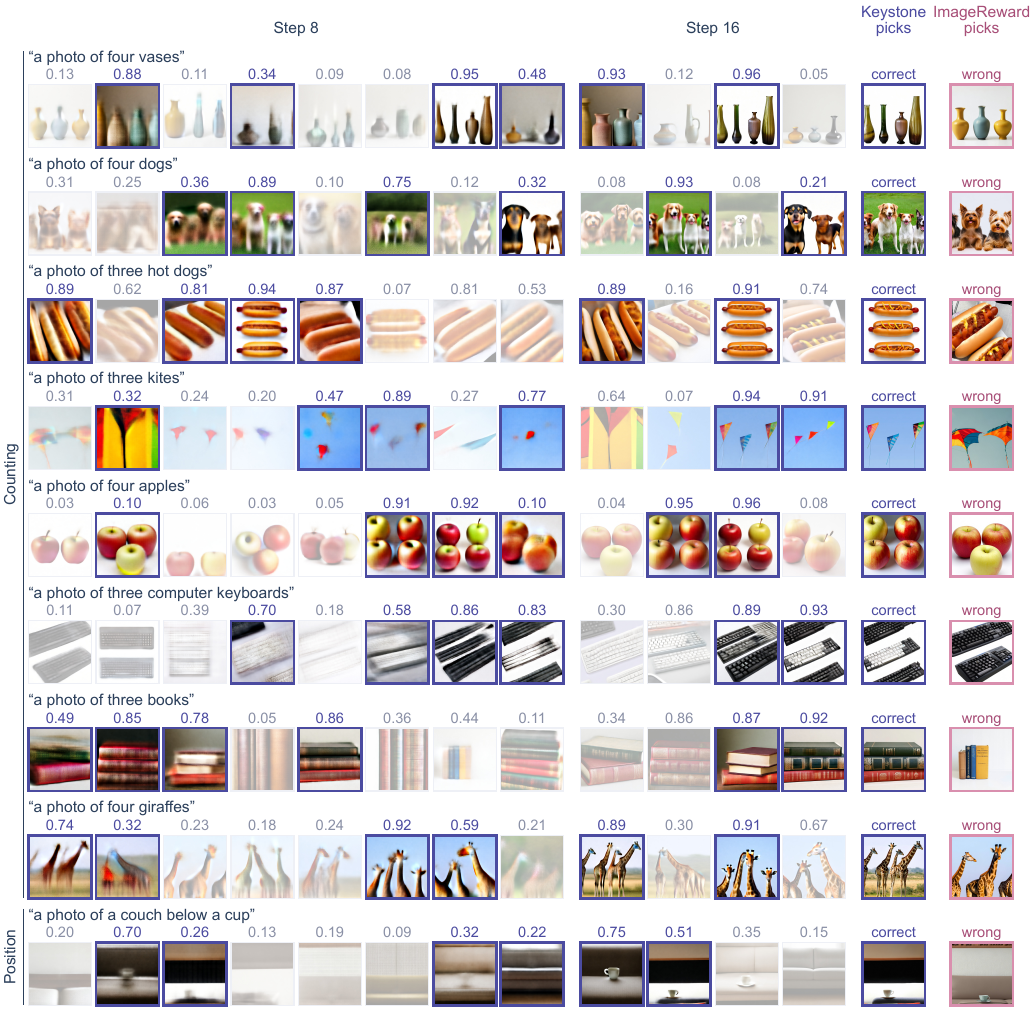}
\caption{\textbf{\method's pruning ends on a correct image where the same pruning driven by ImageReward ends on a wrong one} (SD3.5-Medium, official GenEval test prompts, the candidates of \cref{tab:main}). Each row is one image: its eight step-8 previews with their \method scores (framed: the four kept), the step-16 previews of those four with their scores (framed: the two kept), and the final image each pruning ends on, \method's and that of ImageReward under the same schedule (correct/wrong: verdicts of GenEval's official evaluator); equal scores go to the candidate ranked higher at the previous step, then to the earlier one. Selection: an image qualifies when, under every order of tied scores at every cut, \method's pruning ends on a correct image and ImageReward's on a wrong one; qualifying images per category: counting 47 (in 30 prompts), position 104 (in 62), colour attribution 39 (in 31), two objects 9 (in 7), colours 13 (in 13), single object 1 (in 1); with ties broken in the fixed order of \cref{fig:qual}, the first three categories give \qualWinAll{}. An image is shown only if GPT-5.4 (\cref{sec:app_indep_judge}) agrees: it calls \method's final image a match for the prompt and ImageReward's not. Each prompt contributes its first such image, the prompts of \cref{fig:qual} and \cref{fig:app_qual_more} are left out, and the rows are the first eight prompts of each category by prompt index (or fewer where fewer remain). Continued in \cref{fig:app_gallery_sd35_p2}.}
\label{fig:app_gallery_sd35_p1}
\end{figure}

\begin{figure}[p]
\centering
\includegraphics[width=\linewidth]{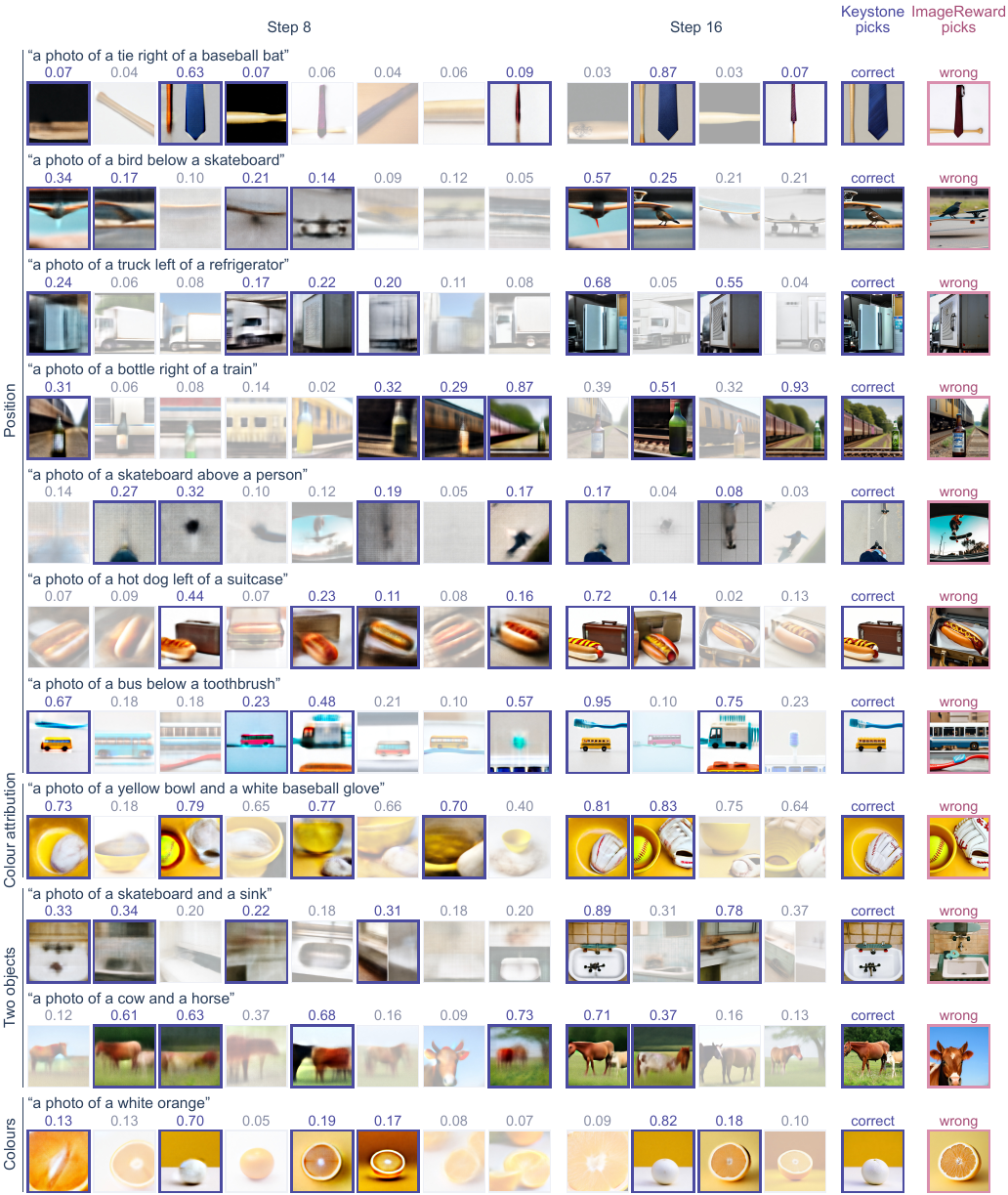}
\caption{\textbf{\method's pruning ends on a correct image where the same pruning driven by ImageReward ends on a wrong one, continued} (position, colour attribution, two objects and colours; setting, layout and selection rule as in \cref{fig:app_gallery_sd35_p1}).}
\label{fig:app_gallery_sd35_p2}
\end{figure}

\begin{figure}[p]
\centering
\includegraphics[width=\linewidth]{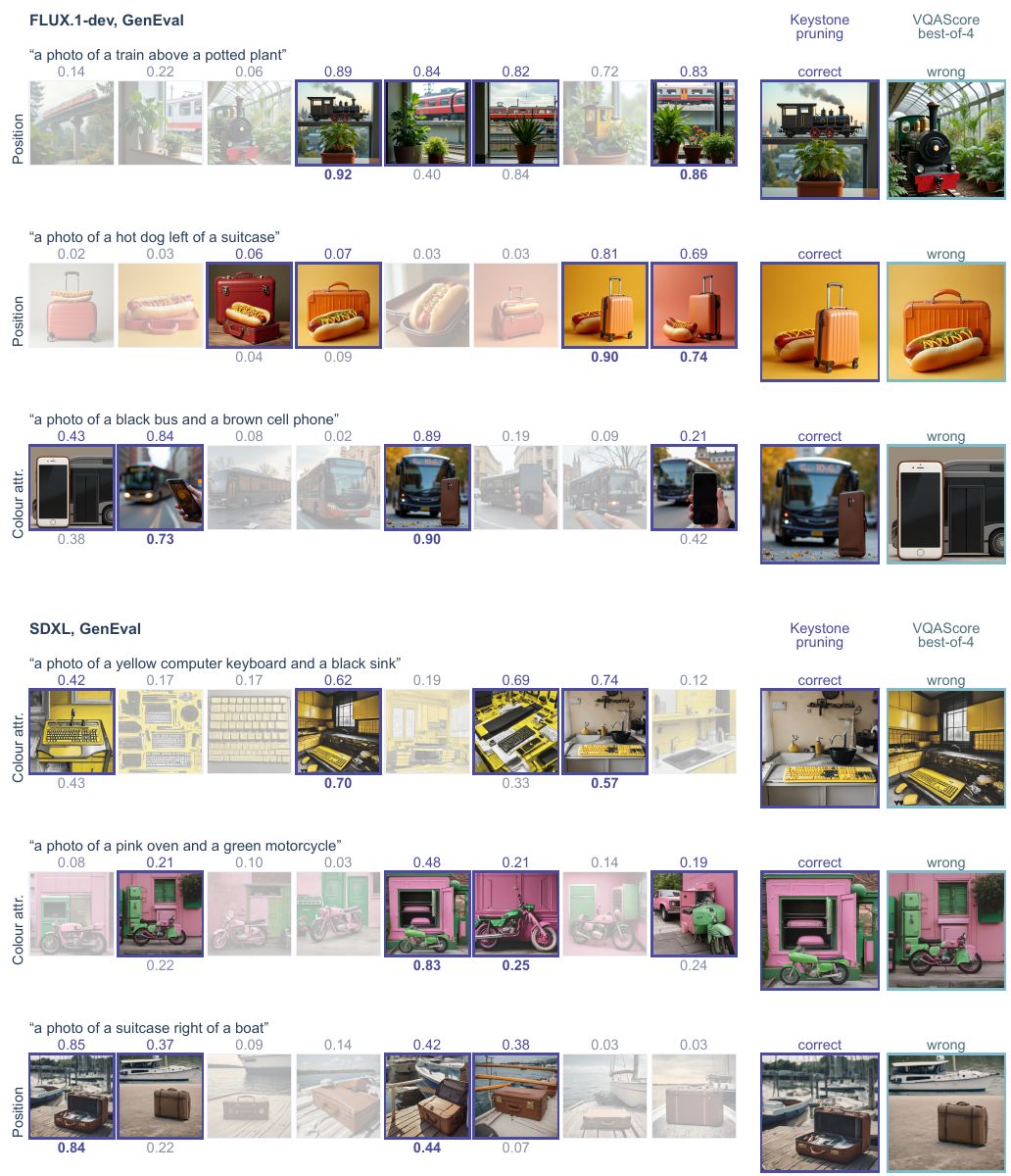}
\caption{\textbf{Examples on FLUX.1-dev and SDXL where \method's pruning ends on a correct image and best-of-4 picks a wrong one} (GenEval, official verdicts). Pruning (144 NFE) against best-of-4 (160 NFE) with the strongest other verifier, VQAScore on both models. Qualifying prompts: 33 for FLUX.1-dev and 58 for SDXL; shown: the first three of each that GPT-5.4 confirms. Thumbnails: finished images, labelled with the \method scores of their step-8 and step-16 previews (layout and rule: \cref{sec:app_gallery_other}).}
\label{fig:app_gallery_other_p1}
\end{figure}

\begin{figure}[p]
\centering
\includegraphics[width=\linewidth]{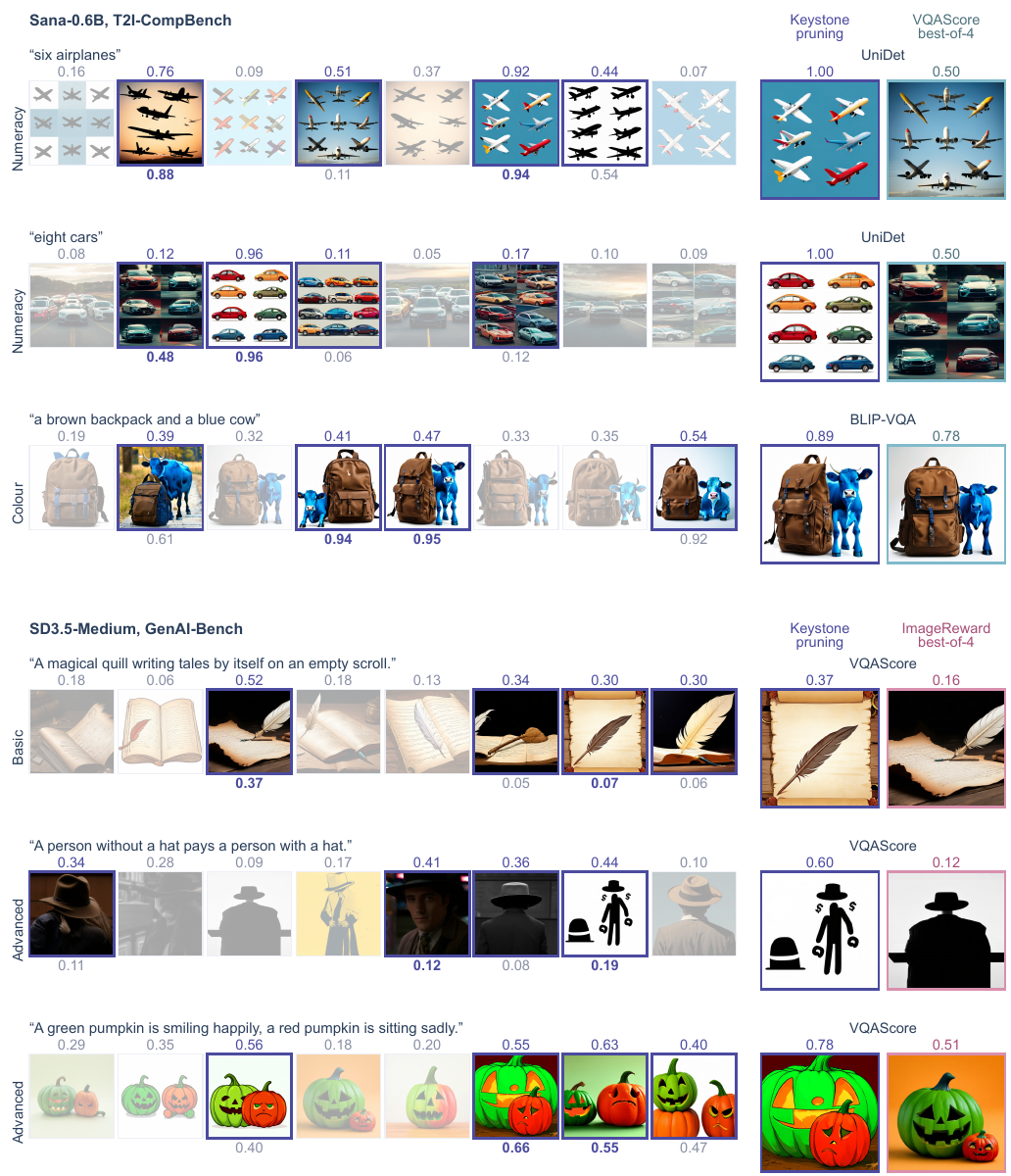}
\caption{\textbf{Examples on Sana-0.6B and SD3.5-Medium where \method's pruning ends on the image the benchmark scores higher}. Pruning (144 NFE) against best-of-4 (160 NFE) with the strongest other verifier. Sana-0.6B on T2I-CompBench: metric BLIP-VQA or UniDet by category, verifier VQAScore, 31 prompts qualify. SD3.5-Medium on GenAI-Bench: metric VQAScore, verifier ImageReward, 21 qualify. Shown: the first three of each that GPT-5.4 confirms; thumbnails (finished images), layout and rule as in \cref{fig:app_gallery_other_p1}.}
\label{fig:app_gallery_other_p2}
\end{figure}

Along the trajectory (\cref{fig:app_mech_sequence}), the \method score separates the candidate that ends correct from the one that ends wrong from step 4 on. On one development prompt (\cref{fig:app_mech_branch}), a new noise draw at the first two steps changes the outcome and no later draw does; over all groups, \cref{fig:keysteps} (left) shows such changes far more often at the first two steps. On identical step-8 previews (\cref{fig:app_mech_misread}), \method's questions prefer the candidate that ends correct where both reward models prefer the wrong one. \Cref{fig:app_gallery_sd35_p1,fig:app_gallery_sd35_p2,fig:app_gallery_other_p1,fig:app_gallery_other_p2} extend \cref{fig:qual} to more categories, base models and benchmarks; each follows the rule stated with it, and from \cref{fig:app_mech_misread} on, GPT-5.4 confirms every outcome shown (\cref{sec:app_indep_judge}).

\clearpage
\subsection{Choice of the Decision Model}
\label{sec:app_decision_model}

\method's questions can be answered by any vision--language model that gives a probability for each answer.
\Cref{tab:app_decision_model} asks the yes/no questions of \cref{sec:app_questions} of Jev, of Qwen3-VL-8B (\cref{sec:app_open_judge}) and of three larger general models served through an API, GPT-5.4, Claude Sonnet 4.6 and Qwen3-VL-235B-A22B, about the same step-8 previews.
Asked the same questions, Jev orders the most candidate pairs correctly; each of its calls costs less than a twentieth of a GPT-5.4 or Claude Sonnet 4.6 call; and it also gives \method's colour choices and boxes, which the yes/no form leaves out.
We therefore use Jev throughout.

\begin{table}[!htbp]
\centering
\footnotesize
\setlength{\tabcolsep}{3.5pt}
\begin{tabular}{lcccccccc}
\toprule
Decision model & Overall & Single & Two obj. & Count. & Colours & Pos. & Col. attr. & \$ / 1k calls \\
\midrule
Jev & \textbf{0.71} & \textbf{0.66} & \textbf{0.73} & \textbf{0.88} & \textbf{0.62} & \textbf{0.77} & \textbf{0.58} & 0.065 \\
GPT-5.4 & 0.67 & 0.57 & 0.69 & 0.81 & \textbf{0.62} & \textbf{0.77} & 0.56 & 1.43 \\
Qwen3-VL-8B & 0.67 & 0.64 & 0.69 & 0.82 & \textbf{0.62} & 0.66 & 0.57 & local \\
Claude Sonnet 4.6 & 0.66$^\ast$ & 0.52 & 0.70 & 0.84 & \textbf{0.62} & 0.71 & 0.55 & 1.83 \\
Qwen3-VL-235B-A22B & 0.64$^\ast$ & 0.61 & 0.67 & 0.77 & 0.59 & 0.69 & 0.53 & 0.27 \\
\midrule
Jev, typed questions (as in the paper) & 0.71 & 0.66 & 0.73 & 0.88 & 0.62 & 0.81 & 0.56 & 0.065 \\
\bottomrule
\end{tabular}

\caption{\textbf{Asked the same questions, Jev orders step-8 previews best, at a fraction of the cost per call} (SD3.5-Medium, official test prompts, the first of the four images per prompt: 553 images of eight candidates each, so the numbers differ from \cref{tab:main}). Pairs ordered correctly at step 8: the share of candidate pairs with different official verdicts that the score orders correctly, ties counted half, per category and averaged over the six. Every model answers the yes/no questions; the last row is Jev with \method's typed questions, as in the paper. Each API model answers all questions about an image in one call, as whole-number probabilities from 0 to 100 (5--20\% of its pairs tie); GPT-5.4 is run at its default sampling; Claude Sonnet 4.6 adds free-text reasoning before its answers in 5\% of replies. Bold: best among the yes/no rows; $^\ast$: below Jev, paired 95\% bootstrap interval of the difference above zero. Cost per 1,000 calls: measured tokens at Bedrock's list price for the API models, the billed cost for Jev; Qwen3-VL-8B runs locally.}
\label{tab:app_decision_model}
\end{table}

\end{document}